\PassOptionsToPackage{table}{xcolor}
\documentclass[manuscript, screen]{acmart}
\AtBeginDocument{%
  }

\setcopyright{acmlicensed}
\copyrightyear{2018}
\acmYear{2018}
\acmDOI{XXXXXXX.XXXXXXX}

\acmJournal{TKDD}
\acmVolume{37}
\acmNumber{4}
\acmArticle{111}
\acmMonth{8}

\usepackage[table]{xcolor}

\usepackage{multirow}
\usepackage[linesnumbered,ruled]{algorithm2e} 

\usepackage[hang,flushmargin]{footmisc}
\usepackage{algpseudocode}
\usepackage{textcomp} 
\usepackage{graphicx}
\usepackage{epstopdf}
\usepackage{rotating}
\usepackage{bbm}
\usepackage{newfloat}
\usepackage{listings}

\usepackage{hhline}
\usepackage{bibentry}
\usepackage{amsmath}
\usepackage{amsthm}

\usepackage{amssymb}
\usepackage{dsfont}
\usepackage{mathrsfs}

\usepackage{mathabx}
\usepackage{booktabs}

\usepackage{enumerate}

\newtheorem{definition}{Definition}
\newtheorem{prop}{Proposition}

\DeclareMathOperator*{\argmin}{argmin}
\DeclareMathOperator*{\argmax}{argmax}

\usepackage{amsfonts}
\usepackage{array}
\usepackage[caption=false,font=normalsize,labelfont=sf,textfont=sf]{subfig}
\usepackage{subcaption}
\usepackage{textcomp}
\usepackage{stfloats}
\usepackage{url}
\usepackage{verbatim}

\begin{document}

\title{Distributed Clustering based on Distributional Kernel}


\author{Hang Zhang}
\affiliation{%
	\institution{
	National Key Laboratory for Novel Software Technology, Nanjing University, China. School of Artificial Intelligence, Nanjing University}
	\country{China}
}
\email{zhanghang@lamda.nju.edu.cn}

\author{Yang Xu}
\affiliation{%
	\institution{
	National Key Laboratory for Novel Software Technology, Nanjing University, China. School of Artificial Intelligence, Nanjing University}
	\country{China}
}
\email{xuyang@lamda.nju.edu.cn}

\thanks{Hang Zhang and Yang Xu contribute equally to this work.}

\author{Lei Gong}
\affiliation{%
	\institution{
	National Key Laboratory for Novel Software Technology, Nanjing University, China. School of Artificial Intelligence, Nanjing University}
	\country{China}
}
\email{gongl@lamda.nju.edu.cn}

\author{Ye Zhu}
\affiliation{
School of Information Technology,  \institution{Deakin University}
  \country{Australia}
}
\email{ye.zhu@ieee.org}

\author{Kai Ming Ting}
\authornote{Corresponding author}
\affiliation{%
	\institution{
	National Key Laboratory for Novel Software Technology, Nanjing University, China. School of Artificial Intelligence, Nanjing University}
	\country{China}
}
\email{tingkm@nju.edu.cn}

% \author{Hang Zhang}

\renewcommand{\shortauthors}{Zhang et al.}

\begin{abstract}
 This paper introduces a new framework for clustering in a distributed network called Distributed Clustering based on Distributional Kernel  ($\mathcal{K}$) or $\mathcal{K}$DC that produces the final clusters based on the similarity with respect to the distributions of initial clusters, as measured by $\mathcal{K}$. It is the only framework that satisfies all three of the following properties. First, $\mathcal{K}$DC guarantees that the combined clustering outcome from all sites is equivalent to the clustering outcome of its centralized counterpart from the combined dataset from all sites. Second, the maximum runtime cost of any site in distributed mode is smaller than the runtime cost in centralized mode. Third, it is designed to discover clusters of arbitrary shapes, sizes and densities. To the best of our knowledge, this is the first distributed clustering framework that employs a distributional kernel. The distribution-based clustering leads directly to significantly better clustering outcomes than existing methods of distributed clustering. In addition, we introduce a new clustering algorithm called Kernel Bounded Cluster Cores, which is the best clustering algorithm applied to $\mathcal{K}$DC among existing clustering algorithms. We also show that $\mathcal{K}$DC is a generic framework that enables a quadratic time clustering algorithm to deal with large datasets that would otherwise be impossible.
\end{abstract}




\received{20 February 2007}
\received[revised]{12 March 2009}
\received[accepted]{5 June 2009}

\maketitle

\section{Introduction}
\label{sec:intro}

Clustering is a fundamental problem in data analysis, and clustering algorithms are widely used in various applications. The rise of big data necessitates a large amount of data to be stored on distributed sites \cite{ordonez2010database,wavecluster,fritz2022efficient,wang2022pack, lu2020distributed}. As a result, data mining tasks often need to be conducted in a distributed framework.

Current work in distributed clustering has focused on converting an existing centralized clustering algorithm into a distributed version. The key challenge is to reduce the high cost of frequent inter-site communication between various sites, while strive to minimize the potential decline in clustering quality \cite{scalablekmeans, RP-DBSCAN, EXDPCTKDD}.

There are two main approaches. The first approach relies on $k$-means clustering \cite{hartigan1979algorithm}. It uses the clustering results of a small representative subset to guide the clustering of the overall data. Specifically, this approach obtains $k$ centers by $k$-means in the subset and then completes the labeling of the entire data through the $k$ centers. 
%An existing framework ($\dddot{\mathbbm{B}}$ in Figure \ref{fig:framework}) of 
This approach requires no parallelization of $k$-means, and its main focus is to obtain a good subset.

The second approach aims to enable a computationally expensive centralized clustering algorithm to deal with large datasets by parallelizing the clustering algorithm. Its focus is to reduce the communication cost and runtime with some approximation techniques that often use indexing such as Locality-Sensitive Hashing (LSH) \cite{lsh04} or kd-tree.
Studies in this approach have focused on density-based clustering algorithms like DBSCAN \cite{DB1996, RP-DBSCAN} and DP \cite{DP2014, EXDPCTKDD, zhang2016efficient} because these algorithms often have better clustering quality than that produced by $k$-means (used in the first approach).

The common characteristic of these two approaches is that each method is intricately crafted for a specific clustering algorithm. 
None of them can be easily adapted to a different clustering algorithm.

In this paper, we propose a new generic framework that is applicable for any existing or new clustering algorithm.

Our work is closely related to the first approach with three steps, but has distinct differences.

\begin{figure}[b]
    \centering
    \includegraphics[width=0.7\textwidth]{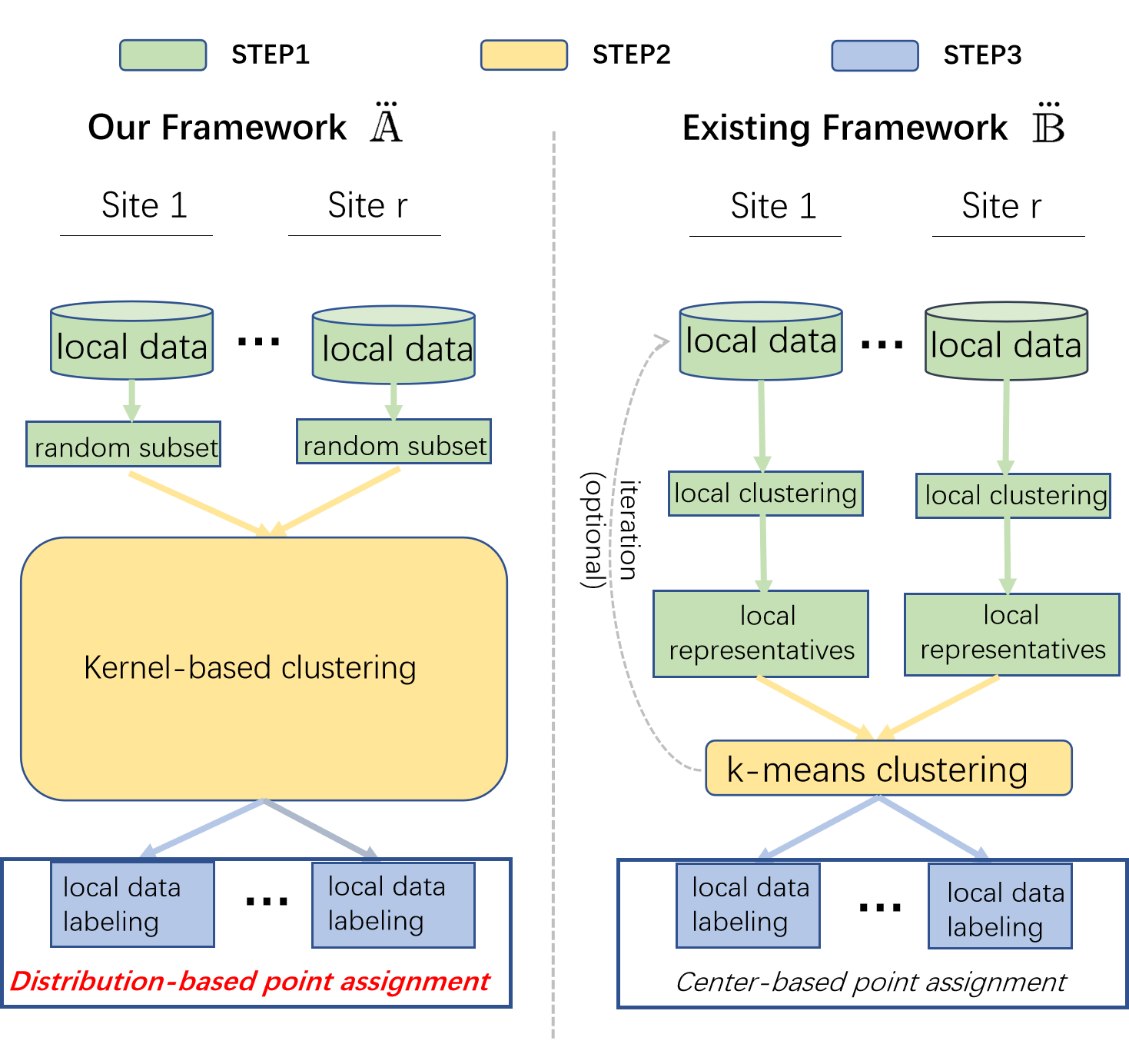}
    \caption{A top-level comparison of distributed clustering: proposed Framework $\dddot{\mathbbm{A}}$ versus existing Framework $\dddot{\mathbbm{B}}$. 
    }
    \label{fig:framework}
\end{figure}

Existing distributed methods under the first approach are based on Framework $\dddot{\mathbbm{B}}$ shown in Figure \ref{fig:framework}(right). 
It is the most direct and efficient way of applying  $k$-means, and it has three steps: 1. Extracting a subset: each distributed site is required to obtain a small set of representative points. This process often involves a clustering algorithm (not necessarily $k$-means) on each site. 2. Clustering: $k$-means on the coordinator site in the network determines the $k$ centers of $k$ clusters from the combined set of representative points from all sites. 3. Point assignment on each local site: the $k$ cluster centers are broadcast from the coordinator site to each local site, and then every point in the local dataset is assigned to the nearest cluster center at each site independently. 

The focus of Framework $\dddot{\mathbbm{B}}$ is to  seek a subset of good representative points (using some method of subset extraction on each site) in step 1, using either one-round communication or multi-round iterative communication, and all existing methods of this framework have used the same last two steps.

Framework $\dddot{\mathbbm{B}}$ has three fundamental limitations. First, all existing methods based on $\dddot{\mathbbm{B}}$ focus on ways to find a `good' representative subset. Yet, even if a good subset of representative points is found, the framework cannot produce a good enough clustering outcome because $k$-means can discover clusters of globular shapes only. Second, 
there is no guarantee that the distributed clustering produces the same clustering outcome as derived by its centralized counterpart. The only exception is the coreset-based methods \cite{Coreset-kmeans-2013,coreset-time} which guarantees that the coreset found in step 1 is as good as the entire dataset. But the first fundamental limitation remains.
Third, $\dddot{\mathbbm{B}}$ often has time complexity worse than linear because of the high computational cost in step 1, despite the use of $k$-means in step 2. 
%although a parallel version of kernel k-means \cite{wang2019scalable} has been proposed for a long time, since the communication between sites cannot be infinitely repeated in distributed mode, kernel $k$-means cannot be directly used in $\dddot{\mathbbm{B}}$.

We are motivated to address these fundamental limitations, especially the first one. First and foremost, we aim to find clusters of arbitrary shapes, sizes and densities, which even existing density-based clustering algorithms such as Density Peak \cite{DP2014} have difficulty finding. We show that this can be achieved via the kernel-based clustering in step 2 and a distribution-based point assignment in step 3 in the framework.

The proposed new  Framework $\dddot{\mathbbm{A}}$ (shown in Figure \ref{fig:framework}(left)) is distinguished from the existing Framework $\dddot{\mathbbm{B}}$ in three aspects:
\begin{enumerate}
    \item The focus of $\dddot{\mathbbm{A}}$ is in steps 2 and 3, and they spend the least amount of time in step 1. In contrast, the most costly component in $\dddot{\mathbbm{B}}$ is in step 1 which is its focus.
    \item It is the first time that distribution-based point assignment is used in distributed clustering. Because clusters are represented as probability density functions, $\dddot{\mathbbm{A}}$ enables final clusters of arbitrary shapes, sizes and densities to be discovered. Although existing Framework $\dddot{\mathbbm{B}}$ may produce clusters of arbitrary shapes in step 2 (by using a non-$k$-means clustering), the center-based point assignment in step 3 limits its final clusters to globular shapes only. 
    \item $\dddot{\mathbbm{A}}$ has linear time complexity. In contrast, both the $k$-means clustering algorithm and  the $k$-means based Framework $\dddot{\mathbbm{B}}$ have super-polynomial time complexity  \cite{k-means-SCG2006}, though they may exhibit linear time on some datasets. This is because (a) $\dddot{\mathbbm{A}}$  executes a clustering algorithm only once in step 2; and (b) $\dddot{\mathbbm{B}}$ usually requires either k-means or some other clustering algorithm to be executed multiple times in step 1. This is the cause of its high computational cost in $\dddot{\mathbbm{B}}$. 
\end{enumerate}

As a result, the proposed distributed clustering Framework $\dddot{\mathbbm{A}}$ has none of the above-mentioned fundamental limitations of $k$-means based Framework $\dddot{\mathbbm{B}}$ for distributed clustering.

In this paper, we make the following contributions:
\begin{itemize}

    \item Proposing a new  framework called Distributed Clustering based on Distributional Kernel ($\mathcal{K}$DC). It is the first distributed-native solution which does not derive from an existing clustering algorithm. The existing framework can be viewed as a degenerated version of $\mathcal{K}$DC which uses $k$-means and center-based point assignment, and it employs no kernel. 

    \item Revealing that $\mathcal{K}$DC has three properties, i.e., (a) $\mathcal{K}$DC and its centralized counterpart are guaranteed to produce the same clustering outcomes, (b) $\mathcal{K}$DC's runtime is guaranteed to be shorter than that of the centralized counterpart, and (c) they both discover clusters of arbitrary-shapes, sizes and densities. Existing methods of $\dddot{\mathbbm{B}}$ satisfy only one out of the three properties.
    \item Creating a new clustering algorithm called Kernel Bounded Cluster Cores ($\kappa$BCC).  We show that $\kappa$BCC is a better candidate than $k$-means and density-based clustering algorithms in step 2 of $\mathcal{K}$DC.

    \item Showing that the proposed centralized counterpart of $\mathcal{K}$DC (that employs $\kappa$BCC) achieves better performance than existing centralized clustering algorithms both in terms of clustering outcomes and runtime efficiency. 

\end{itemize}

The key notations used are shown in Table \ref{notations}.

\begin{table}
\centering
\caption{Key notations used.}
\label{notations}
\begin{tabular}{@{}c|l@{}}
\hline
% & Definition \\ \hline
$\kappa$ & Point-to-point kernel $\kappa(\mathbf{x},\mathbf{y})$ \& $\mathbf{x},\mathbf{y} \in \mathbb{R}^d$\\
$\mathcal{K}$ & Distributional kernel $\mathcal{K}(\mathcal{P},\mathcal{Q})$ \& $\mathcal{P},\mathcal{Q}$ are pdfs in $\mathbb{R}^d$\\
$k$ & The number of clusters \\ 
$r$ & The number of local sites in a network \\
%$n$ & Total number of instances from all sites \\
%$d$ & Dimension of the data \\
%$s$ & Subset data size $s = |at the end of step 1 \\
$\mathbbm{A}$ & Centralized clustering framework based on $\mathcal{K}$\\
$\dddot{\mathbbm{A}}$ & Distributed clustering framework based on $\mathcal{K}$\\
$\dddot{\mathbbm{B}}$ &  Distributed clustering framework based on $k$-means\\
$\mathbbm{f}$ & clustering algorithm used in step 2 in $\mathbbm{A}$ or $\dddot{\mathbbm{A}}$\\
$\mathcal{B}$ & Set of representative points from all sites \& $s=|\mathcal{B}|$\\
$\mathcal{B}_\ell$ & Subset of representative points on site $\ell$\\
$D$ & Combined dataset from all sites, where $n=|D|$\\
$D_\ell$ & Dataset on site $\ell$\\
%$\mathcal{G}_i$ & Cluster $i$\\
%$C_{j}$  & Labels for each point in the final clustering result \\
\hline
\end{tabular}
\end{table}

\section{Related Work}
%{Current approaches for distributed clustering}
\label{sec_survey}

Here we present two current approaches for distributed clustering.
Three types of methods in the existing Framework $\dddot{\mathbbm{B}}$ of the first approach of distributed clustering are described in Section \ref{sec-first-approach}, and the second approach is presented in Section \ref{sec-second-approach}.

\begin{table*}[!htb]
\caption{Characteristics of distributed clustering: the proposed Framework $\dddot{\mathbbm{A}}$ versus existing Framework $\dddot{\mathbbm{B}}$ of different methods.\\
      $\dagger$ Step 3 has time complexity $O(n)$, and step 2 has time complexity $O(s^2)$, where $s \ll n$ for a large dataset.}
\label{compa-alg}
 \setlength{\tabcolsep}{1pt}
 \scalebox{0.75}{
\centering
      \begin{tabular}{c|c|l|c|c|c|c|c|c|c}
      \hline
\multicolumn{2}{l|}{}   &       & {Time } & \multicolumn{3}{c|}{Communication cost  } & \multicolumn{3}{c}{Final clustering quality}  \\\cline{4-10} 
\multicolumn{2}{l|}{Framework \& Step-1-Type} & Example methods       & {Most costly} & {\multirow{2}[1]{*}{Constant}} & {One-round } & {No communication } & {Varied densities} & {Consistent with } & {Arbitrary} \\
\multicolumn{2}{l|}{}   &  & {step} & {} & {communication} & {between sites} & {and sizes} & {the combined set} & {shapes} \\ \hline
$\dddot{\mathbbm{A}}$ &  Random-sample-based & $\mathcal{K}$DC (Ours)  & steps 2 \& 3$^\dagger$  & \checkmark  & \checkmark  & \checkmark  & \checkmark  & \checkmark  & \checkmark \\ \hline
\multirow{8}{*}{$\dddot{\mathbbm{B}}$} &    \multirow{3}{*}{Density-based} & DBDC \cite{DBDC-2004}  &  step 1    & \cellcolor{gray!50}  & \checkmark  & \checkmark  & \checkmark   & \cellcolor{gray!50}      & \cellcolor{gray!50} \\
    &        & S-DBDC \cite{SDBDC-2004} &  step 1     & \cellcolor{gray!50}  & \checkmark  & \checkmark  & \checkmark   & \cellcolor{gray!50}      & \cellcolor{gray!50} \\
     &       & L-DBDC \cite{LDBDC-2008} &  step 1     & \cellcolor{gray!50}  & \checkmark  & \checkmark  & \checkmark  & \cellcolor{gray!50}      & \cellcolor{gray!50} \\
      \cline{2-10}
    % \hhline{*{2}{|~}*{9}{|-}|~}
    &  \multirow{3}{*}{Iteration-based} & MR-$k$-Center \cite{Map-kCenter-2011} &   step 1  & \cellcolor{gray!50}  &  \cellcolor{gray!50}     & \checkmark  &  \cellcolor{gray!50}     &  \cellcolor{gray!50}     & \cellcolor{gray!50} \\ 
     &        & CODC \cite{CODC-2016}  & step 1  & \cellcolor{gray!50}  & \cellcolor{gray!50}      & \checkmark  &  \cellcolor{gray!50}     & \cellcolor{gray!50}      & \cellcolor{gray!50} \\
      &      & PLSH \cite{lsh18}  & step 1  & \cellcolor{gray!50}  & \cellcolor{gray!50}      & \checkmark  &  \cellcolor{gray!50}     & \cellcolor{gray!50}      & \cellcolor{gray!50} \\
        \cline{2-10}
       % \hhline{*{1}{|~}*{9}{|-}|~}
&     \multirow{2}{*}{Coreset-based} & $k$-means \cite{NIPS2013_Balcan,NEURIPS2018_Li} & step 1  & \checkmark  & \checkmark  & \checkmark  &  \cellcolor{gray!50}     & \checkmark  & \cellcolor{gray!50} \\
  &      & DR-$k$-means \cite{DR-kmeans-2015,ipca14} & step 1      & \checkmark  & \checkmark  & \checkmark  &   \cellcolor{gray!50}    & \checkmark  & \cellcolor{gray!50} \\ \hline
      \end{tabular}}
\end{table*}%

\begin{table*}[]
\caption{The three steps in the proposed Framework $\dddot{\mathbbm{A}}$ and existing Framework $\dddot{\mathbbm{B}}$. $\bar{\mathbf{x}}_{\mathcal{G}_i}= \frac{1}{|\mathcal{G}_i|} \sum_{\mathbf{y} \in \mathcal{G}_i} \mathbf{y}$. 
     Density-based methods employ DBSCAN  \cite{DB1996} to produce a subset $\mathcal{B}$ of high-density points only; the subsequent clustering on $\mathcal{B}$ is performed using $k$-means. Note that, unlike other existing methods, the proposed random-sample-based method uses no clustering algorithms to produce $\mathcal{B}$.\\
     $\ddagger$ While Framework $\dddot{\mathbbm{A}}$ admits any clustering algorithm, the actual method used may impact on this property. See Section \ref{sec-property-c} for details. }
    \label{tab:compare}
   \scalebox{0.85}{
    \centering
    \begin{tabular}{c|c|c|c|c|c|c|c}
    \hline
\multicolumn{2}{l|}{Framework \& Step-1-Type} & \multicolumn{3}{c|}{Distributed Clustering} & \multicolumn{3}{c}{$\mathcal{K}$DC properties}\\\cline{3-8} 
    \multicolumn{2}{c|}{} & Step 1: $\mathcal{B}$ from $\bigcup_\ell D_\ell$ & Step 2: Clustering on $\mathcal{B}$ & Step 3: Assign $\mathbf{x} \in D_\ell$ to & (a) & (b) & (c)\\ \hline
   \multirow{2}[2]{*}{$\dddot{\mathbbm{A}}$} & Random-sample-based & Random subset  & $\mathbbm{f}(\mathcal{B}) \rightarrow k$ clusters $\mathcal{G}_i$  & %$\mathop{\argmin}\limits_{i \in [1,k]} \Vert \phi(\mathbf{x}) - \widehat{\phi}(\mathcal{G}_i) \Vert$
   $\mathop{\argmax}\limits_{i \in [1,k]} \mathcal{K}(\delta(\mathbf{x}),\mathcal{P}_{C_i})$ &
   %\left< \Phi(\delta(\mathbf{x})),{\Phi}(\mathcal{P}_{\mathcal{G}_i}) \right>$ & 
   $\checkmark$ & $\checkmark$  & $\checkmark^\ddagger$ \\
   & & [sampling] & $\mathbbm{f} =$ any clustering & & &  &  \\ \hline  \multirow{7}[2]{*}{$\dddot{\mathbbm{B}}$} & Density-based & High-density subset& &  &$\times$ &$\times$ &$\times$\\
   & & [DBSCAN] & & & &\\ \cline{2-3} \cline{6-8}
   & Iteration-based  & Subset after multiple iterations & $\mathbbm{f}(\mathcal{B}) \rightarrow k$ clusters $\mathcal{G}_i$ & $\mathop{\argmin}\limits_{i \in [1,k]} \Vert \mathbf{x} - \bar{\mathbf{x}}_{\mathcal{G}_i} \Vert$ &$\times$ &$\times$ &$\times$\\
   & & [$k$-means] & $\mathbbm{f} =$ $k$-means &  & &\\ \cline{2-3} \cline{6-8}
   & Coreset-based   & Coreset &  & & $\checkmark$ & $\times$ &$\times$\\
   & & [$k$-means]  &   &   &  & & \\ \hline

%     &  & $\mathbbm{f} =$ $k$-means & & $\checkmark$ & $\checkmark$  & $\times$ \\
%     &  & $\mathbbm{f} =$ $\kappa$BCC & & $\checkmark$ & $\checkmark$  & $\checkmark$ \\ 
    \end{tabular}}
\end{table*}

\subsection{First Existing Approach: Framework $\dddot{\mathbbm{B}}$}
\label{sec-first-approach}

Framework $\dddot{\mathbbm{B}}$ tailors to $k$-means clustering without parallelization, focusing on finding good local representative samples at each site. 

\textbf{Density-based algorithms}. To improve the original purely $k$-means based clustering outcomes, some density-based methods are proposed to obtain subsets of better representative points  \cite{DBDC-2004,SDBDC-2004,LDBDC-2008} for $k$-means. The idea is to use DBSCAN to produce a subset that consists of clusters of arbitrary shapes and sizes at each site in step 1 in Framework $\dddot{\mathbbm{B}}$.

One disadvantage of using DBSCAN is that, due to the differences of data distributions at various sites, it is difficult to find a single set of parameter settings for DBSCAN that are suitable at all sites \cite{LDBDC-2008}.

\textbf{Iteration-based algorithms} perform $k$-means after sampling at each iteration in step 1. 
% \textcolor{red}{It is unclear you are referring to step 1 or 2, as $k$-means is applied in both steps.}

The main point is that the clustering results brought by the subset obtained by random sampling only once may be unreliable. Therefore, these methods \cite{Map-kCenter-2011,CODC-2016,lsh18} retain $k$ centers after the weighted $k$-means and add them to the next-sampled subset to calculate a new round of $k$-means results.

When the difference between the $k$ centers obtained in two consecutive rounds is below a threshold, the algorithm stops. Afterwards, the $k$ centers will be communicated to each site to complete the label assignment.

The design pattern of this algorithm leads to a necessarily multi-round communication model, which greatly increases the communication cost and runtime.

\textbf{Coreset-based algorithms} \cite{NIPS2013_Balcan,NEURIPS2018_Li,DR-kmeans-2015,ipca14} are the only one in $\dddot{\mathbbm{B}}$ with guaranteed clustering results.

The idea is to calculate a small weighted subset to ensure that the $k$ centers obtained by $k$-means on this subset are the same as the $k$ centers obtained by $k$-means with the entire dataset.

Coreset-based $k$-means \cite{NIPS2013_Balcan} gives an implementation of coreset-based $k$-means in a distributed framework. Subsequent work DR-$k$-means \cite{DR-kmeans-2015,ipca14} implements its dimensionality-reduced version to reduce the communication cost, and dist-kzc \cite{NEURIPS2018_Li} implements a version adapted to noisy data.

Coreset-based $k$-means is the state-of-the-art (SOTA) algorithm in $\dddot{\mathbbm{B}}$, which theoretically guarantees the same clustering results as the centralized $k$-means. 

The other two types of methods mentioned above in $\dddot{\mathbbm{B}}$ can only approximate the clustering outcomes of the centralized $k$-means. 

% \vspace{3mm}
% \newpage
\noindent
\textbf{Summary: Fundamental limitation of Framework $\dddot{\mathbbm{B}}$ }

The key fundamental limitation of Framework $\dddot{\mathbbm{B}}$ is that it uses center-based point assignment in step 3. As a result, irrespective of the quality of the subset obtained in step 2, $k$-means produces poor clustering outcomes on many real datasets which have clusters of non-globular shapes and non-equal sizes and densities (see Section \ref{sec-property-c} for details).

The characteristics of methods in Framework $\dddot{\mathbbm{B}}$, in comparison to those in the proposed Framework $\dddot{\mathbbm{A}}$, are given in Tables \ref{compa-alg} and \ref{tab:compare}.

\subsection{Second Existing Approach}
\label{sec-second-approach}

The second approach focuses on parallelization of a clustering algorithm by reducing communication cost and improving time efficiency  through indexing techniques and stochastic strategies.
An early example is parallel $k$-means  \cite{parakmeans2000}.

LDSDC \cite{lds20} proposes a distributed clustering algorithm based on Gaussian Mixture Models (GMM) \cite{gmm2015}. It performs density-based clustering followed by Gaussian modeling at each site. The model parameters at each site are then transmitted to the coordinator to produce a global Gaussian mixture model. The local clusters at each site are then adjusted in accordance with the global model. Its main problem is that the clustering quality is easily affected by the number of sites, and is also sensitive to its hyperparameter settings \cite{lds20}.

By leveraging random partitioning techniques, RP-DBCSAN \cite{RP-DBSCAN} provides a superfast, parallelized version of DBSCAN, addressing the challenges of efficiency and scalability in handling large datasets.
%while maintaining high-quality clustering results.

LSH-DDP \cite{ddp16} is a distributed version of the Density-Peak \cite{DP2014} algorithm under the MapReduce infrastructure. It breaks down the distance calculations and density estimations required for all points in a given dataset into five MapReduce jobs, and a centralized procedure for density peak selection and point assignment. The mapping procedure utilizes Locality-Sensitive Hashing (LSH) \cite{lsh04} in order to facilitate computational speedup. %This trades off the quality of clustering outcome with efficiency, though it enables large datasets to be processed.

Instead of LSH, Ex-DPC++ \cite{EXDPCTKDD, EXDPCSIGMOD} uses a box-based density estimator (instead of $\epsilon$-neihgborhood density esitmator used in DP). This allows it to employ kd-tree to obtain the points in a box and use cover-tree to accelerate nearest neighbor search. This change enables range counting on a kd-tree, facilitating concurrent computation across multiple threads. 

A common key issue with the second approach is that all methods trade off clustering quality with efficiency to enable a target clustering algorithm to run on large datasets. It is interesting to note that most of these works do not compare the clustering quality of the proposed distributed clustering with that of the original clustering algorithm (e.g., \cite{EXDPCTKDD,lds20, TEPDBSCAN,NG_DBSCAN}). This clustering quality gap must be ascertain in persuit of parallelization. A fast parallelization is of no use if it produces poor clustering quality.

The proposed Framework $\dddot{\mathbbm{A}}$ is distinguished from the second approach in one key aspect. Framework $\dddot{\mathbbm{A}}$ relies on a good clustering algorithm in step 2 which is able to produce a good set of representative clusters using random data subsets from all sites. No parallelization of the algorithm is required. In contrast, the second approach relies substantially on some indexing schemes, such as LSH and kd-tree,  to parallelize a clustering algorithm.

Given the close relationship between Frameworks $\dddot{\mathbbm{A}}$ and $\dddot{\mathbbm{B}}$, we focus on constrasting them in the main experiment. In addition, we have included a recent method of the second approach, i.e., Ex-DPC++, as a competitor in Section \ref{sec:Ex-dpc++}.

\section{Distributional Kernel-based Clustering}
Distributional kerne-based clustering represents each cluster as a probability density function, which means the points in the same cluster are independent and identically distributed.
Given a dataset $D$,
let $\mathcal{P}_{C}$ be the probability density function (pdf) of a cluster $C \subset D$. 

A distributional kernel $\mathcal{K}(\mathcal{P}_{C_i},\mathcal{P}_{C_j}) = \left< {\Phi}(\mathcal{P}_{C_i}),{\Phi}(\mathcal{P}_{C_j}) \right>$ measures the similarity between two pdfs $\mathcal{P}_{C_i}$ and $\mathcal{P}_{C_j}$, where  ${\Phi}(\mathcal{P})$ is the feature map of the kernel $\mathcal{K}$ that maps the pdf $\mathcal{P}$ in input space into a point $\Phi(\mathcal{P})$ in the feature space of $\mathcal{K}$ \cite{KernelMeanEmbedding2017}.

\begin{definition}
\label{def_clusters} 
Boundaries between probability density functions of clusters.
If dataset $D$ has a set of clusters $\mathbb{C} = \left\{ C_1, \dots, C_k \right\}$,  the boundary between any two of the $k$ clusters is defined for $\mathbf{x} \in \mathbb{R}^d$  as follows:
\begin{align*}
  \mathcal{K}(\delta(\mathbf{x}),\mathcal{P}_{C_i})  =  \mathcal{K}(\delta(\mathbf{x}),\mathcal{P}_{C_j})\ \forall i \ne j \Leftrightarrow   \left< \Phi(\delta(\mathbf{x})),{\Phi}(\mathcal{P}_{C_i}) \right>  =  \left< \Phi(\delta(\mathbf{x})),{\Phi}(\mathcal{P}_{C_j}) \right>\ \forall i \ne j
\end{align*}
\noindent
where    $\delta (\mathbf{x})$ is the Dirac measure of a point $\mathbf{x}$ which converts a point into a pdf.

\end{definition}

%the kernel mean map of $K$ is $\widehat{\phi}(\mathcal{P}_{C}) = \frac{1}{|C|} \sum_{\mathbf{y} \in C} \phi(\mathbf{y})$; and $\phi(\cdot)$ is the feature map of a point-to-point kernel $\kappa$; $\widehat{\phi}(\delta(\mathbf{x})) = \phi(\mathbf{x})$; and

\begin{figure}[t]
    \centering
    \includegraphics[width=0.8\textwidth]{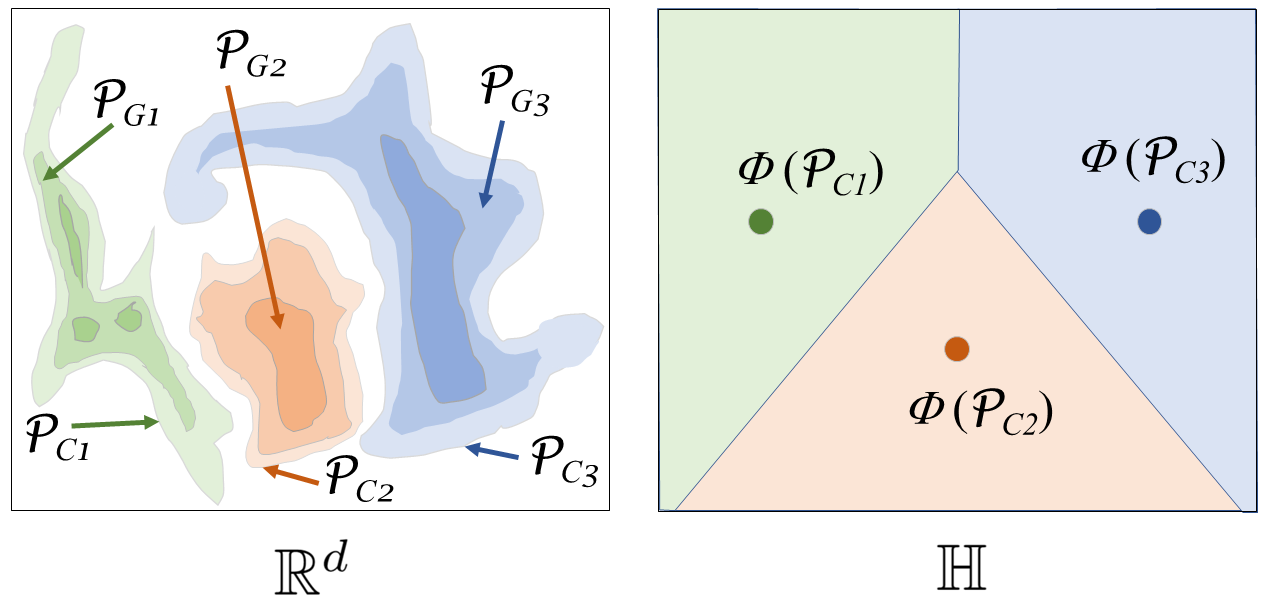}
    \caption{Clusters in the input space $\mathbb{R}^d$ (left), and cluster representations in the feature space $\mathbb{H}$ of distributional kernel $\mathcal{K}$ (right). Clusters $C$ as distributions $\mathcal{P}_C$ in $\mathbb{R}^d$, and as $\Phi(\mathcal{P}_C)$ in $\mathbb{H}$. Initial clusters $\mathcal{G} \subset C$ are identified in Framework $\mathbbm{A}$, given in Section \ref{sec-FrameworkA}.}
    \label{fig:illus}
\end{figure}

In other words, in the feature space of $\mathcal{K}$, the clusters form a Voronoi diagram with a Vonoroi cell centered at ${\Phi}(\mathcal{P}_{C_i})$---representing cluster $C_i$ in the input space---having the boundaries as stated in Definition \ref{def_clusters}. An illustration is shown in Figure \ref{fig:illus}.

The primary advantage of representing each cluster as a pdf is that the clusters can be any arbitrary shapes, sizes and densities. 

\section{Clustering based on Distributional Kernel $\mathcal{K}$}
%{A framework of centralized clustering based on $\mathcal{K}$ that is amenable for distributed clustering}
\label{sec-FrameworkA}
In this section, we provide a new clustering that is native to both centralized and distributed clusterings.
The above definition prompts us to design a centralized clustering framework that has three steps. Step 1 simply samples a subset $\mathcal{B}$ from a given dataset $D$. Step 2 finds some initial clusters ${\mathcal{G}_i} \subset \mathcal{B}$ which are good representatives of the true clusters, i.e., ${\mathcal{G}_i} \subset C_i,\ \forall i$. Step 3 assigns each point in $D$ to its most similar ${\mathcal{G}_i}$, as measured by the distributional kernel $\mathcal{K}$, following Definition \ref{def_clusters}.  
%\[ C_{j}=\{\mathbf{x}\in S\ |\ \mathop{\argmax}\limits_{i \in [1,k]} \left< \Phi(\delta(\mathbf{x})),{\Phi}(\mathcal{P}_{\mathcal{G}_i}) \right> =j \} \]

Framework $\mathbbm{A}$ provides the details of the procedure.

\begin{algorithm}[h]
\renewcommand{\algorithmcfname}{Framework $\mathbbm{A}$}
% \RemoveAlgoNumber
\renewcommand{\thealgocf}{}
\SetAlgoLined
\SetKwInOut{Input}{Input}\SetKwInOut{Output}{Output}
\Input{$D$ - dataset, $k$ - number of clusters,
distributional  kernel $\mathcal{K}(\cdot,\cdot) = \left< \Phi(\cdot),\Phi(\cdot)\right>$.}
\Output{$\mathbb{C} = \{ C_{1}, \dots, C_{k} \}$}
\BlankLine
Get a subset $\mathcal{B} \subset D$\;
Produce $k$ initial clusters $\mathcal{G}_i$ by performing clustering $\mathbbm{f}$ on $\mathcal{B}$, and compute $\Phi(\mathcal{P}_{\mathcal{G}_i})$\;
$\mbox{For } j =1,\dots,k:$ \hspace{3cm}
$C_{j}=\left\{\mathbf{x}\in D\ |\ \mathop{\argmax}\limits_{i \in [1,k]} \left< \Phi(\delta(\mathbf{x})),{\Phi}(\mathcal{P}_{\mathcal{G}_i}) \right> =j \right\}$\;
%$C_{j}=\{\mathbf{x}\in S\ |\ \mathop{\argmin}\limits_{i \in [1,k]} \parallel \phi(\mathbf{x}) - \widehat{\phi}(\mathcal{G}_i) \parallel =j \}$\;
\Return $\mathbb{C} = \{ C_{1}, \dots, C_{k} \}$\;
\caption{Centralized Clustering based on $\mathcal{K}$}
%\label{alg:KClustering}
\end{algorithm}

Note that $\mathbbm{f}$ employed in step 2 can be any clustering algorithm that produces $k$ initial clusters $\mathcal{G}_i, i=1,\dots,k$ from $\mathcal{B}$.

\begin{algorithm}[t]
\renewcommand{\algorithmcfname}{Framework $\dddot{\mathbbm{A}}$}
% \RemoveAlgoNumber
\renewcommand{\thealgocf}{}
\SetAlgoLined
\SetKwInOut{Input}{Input}\SetKwInOut{Output}{Output}
\Input{$D_\ell$ - dataset on site $\ell\ \forall_\ell$ in a network, $k$ - number of clusters,  
%$s$ - sample size,
% $\tau$ - similarity threshold}
distributional kernel  $\mathcal{K}(\cdot,\cdot) = \left< \Phi(\cdot),\Phi(\cdot)\right>$.}
\Output{$\mathbb{C}^\ell = \{ C^{\ell}_1, \dots, C^{\ell}_k \}$  on every site $\ell$}
\BlankLine
On site $\ell\ \forall_\ell$: Get a subset: $\mathcal{B}_\ell \subset D_\ell$ \hspace{3cm}
- Transmit $\mathcal{B}_\ell$ to the coordinator, and $\mathcal{B} = \bigcup_\ell \mathcal{B}_\ell$\;
On the coordinator, produce $k$ initial clusters $\mathcal{G}_i$ by performing clustering $\mathbbm{f}$ on $\mathcal{B}$, and compute ${\Phi}(\mathcal{P}_{\mathcal{G}_i})$ \hspace{3cm}
- Transmit ${\Phi}(\mathcal{P}_{\mathcal{G}_i})\ \forall_i$ to every site $\ell$\;
On site $\ell\ \forall_\ell$:\ 
$\mbox{For } j =1,\dots,k:$ \hspace{3cm} 
$C^{\ell}_j=\left\{\mathbf{x}\in D_\ell\ |\ \mathop{\argmax}\limits_{i \in [1,k]} \left< \Phi(\delta(\mathbf{x})),{\Phi}(\mathcal{P}_{\mathcal{G}_i}) \right> =j \right\}$\;
%$C^{\ell}_j=\{\mathbf{x}\in D_\ell\ |\ \mathop{\argmin}\limits_{i \in [1,k]} \parallel \phi(\mathbf{x}) - \widehat{\phi}(\mathcal{G}_i) \parallel =j \}$\;
\Return $\mathbb{C}^\ell = \{ C^{\ell}_1, \dots, C^{\ell}_k \}$ on every site $\ell$\;
\caption{Distributed Clustering based on $\mathcal{K}$ ($\mathcal{K}$DC)}
%\label{alg:DKClustering}
\end{algorithm}

Two key aspects of Framework $\mathbbm{A}$ are in the last two steps. Step 2: Initial clusters $\mathcal{G}_i$, which are  good representatives of the true clusters, can be produced from a clustering algorithm using a data subset $\mathcal{B} \subset D$. See an example illustration in Figure \ref{fig:illus}, where $\mathcal{G}_i$ is a subset of the intended cluster $C_i$ to be discovered. 
%the clustering is performed on a subset from random sampling, and it can be conducted much faster than that on the whole given dataset.

Step 3: The final point assignment, expanding $\mathcal{G}_i$ to $C_i$, is completed on the entire dataset in one iteration. This is much faster than the typical optimization method such as $k$-means which requires multiple iterations.

We show here that 
%Framework $\mathbbm{A}$ is easily amenable for distributed clustering 
the above clustering is also native to distributed clustering. The detailed procedure is shown in Framework $\dddot{\mathbbm{A}}$, named $\mathcal{K}$DC, where the same three steps are applied. Clustering algorithm $\mathbbm{f}$ is performed once only on a coordinator in step 2 using $\mathcal{B}$; and the final point assignment in step 3 is completed on individual sites in a network.

Other differences between Frameworks $\mathbbm{A}$ and $\dddot{\mathbbm{A}}$ are:
(a) $D$ versus $\bigcup_\ell D_\ell$ in step 1, where $D_\ell$ is a dataset on site $\ell$ in a network; and (b) $D$ versus $D_\ell$ in step 3 in order to produce a clustering outcome on each site $\ell$.

Note that the clustering outcome of the centralized version, $C_i$ is the union of all clustering outcomes of its distributed counterpart $C_i^\ell$ from all sites $\ell$ in the network.

Framework $\dddot{\mathbbm{A}}$ is a distributed-native solution because it is not derived from any existing clustering algorithm, and yet, any existing clustering algorithm can be used as $\mathbbm{f}$ in $\dddot{\mathbbm{A}}$.

We are not aware that a distributed-native solution exists in the literature. Framework $\dddot{\mathbbm{A}}$ can be viewed as the first distributed-native solution and its properties are presented in the next section.

\section{Properties of $\mathcal{K}$DC} 
\label{properties}
We provide the three properties of $\mathcal{K}$DC in this section.

The three properties of $\mathcal{K}$DC are described formally as follows:

\begin{definition}
Distributed Clustering Framework  $\dddot{\mathbbm{A}}$ and its centralized counterpart $\mathbbm{A}$, with any clustering algorithm $\mathbbm{f}$, have the following three properties, irrespective of the data sizes and cluster distributions on different sites in a distributed network of $r$ sites and a coordinator:
\begin{itemize}
    % \item[(a)] The clustering method $\mathbbm{f}$ is executed only once,  independent of the number of sites $r$ in a distributed network; and no additional clustering methods are required to be run in $\mathbbm{A}$ or $\dddot{\mathbbm{A}}$.
    \item[(a)] Distributed-Centralized Clustering Equivalence:\\ \[\mathbbm{A}\left(\bigcup_{\ell=1,\dots,r} D_\ell\right) \equiv\    \bigcup_{\ell=1,\dots,r} \dddot{\mathbbm{A}}(D_\ell). \]
    where $D_\ell$ are individual datasets  on site $\ell$,  the number of clusters in $\bigcup_\ell D_\ell$ is $k$, and the number of clusters in $D_\ell$ is $k_\ell \in [1,k]$.
    %$\mathbbm{A}(\bigcup_\ell D_\ell) \equiv\    \bigcup_\ell\ \dddot{\mathbbm{A}}(D_\ell)$; and 
    \item[(b)] Distributed clustering always runs faster than Centralized clustering:\\ 
    \[\Lambda\left(\mathbbm{A}\left(\bigcup_{\ell=1,\dots,r} D_\ell\right)\right) > \max_{\ell=1,\dots,r} \Lambda\left(\  \dddot{\mathbbm{A}}(D_\ell)\right). \]
    \noindent where $\Lambda(\mathbbm{A})$ is the runtime cost of executing  $\mathbbm{A}$.
    \item[(c)] Clustering performance guarantee:
    \[\mbox{If }\ \mathcal{P}_{\mathcal{G}_i} \approx \mathcal{P}_{\mathcal{T}_i}, \mbox{ then } \mathcal{P}_{\mathcal{C}_i} \approx \mathcal{P}_{\mathcal{T}_i}\ \forall i,\] 
    where $\mathcal{G}_i$ is the initial cluster produced in step 2, $\mathcal{C}_i$ is the final clustering outcome of either of the two proposed frameworks, $\mathcal{T}_i$ is the corresponding ground truth cluster in the given dataset, and $\mathcal{G}_i \subset \mathcal{C}_i \subset D$.
\end{itemize}
\label{def_DKC}
\end{definition}

Note that $\max_\ell \Lambda(\dddot{\mathbbm{A}}(D_\ell, \ell=1,\dots,r))$ includes the overhead cost such as the communication cost between sites, and other local processing time required 
%(possibly running additional $\mathbbm{f}$ or other procedures locally 
on individual sites.

As property (c) is based on pdf, it admits clusters of arbitrary shapes, sizes and densities. As a result, it enables $\mathcal{K}$DC to produce better clustering quality than those derived from frameworks based on $k$-means as well as many existing centralized clustering algorithms.

These three properties make $\mathcal{K}$DC a better candidate for distributed clustering tasks than existing methods of distributed clustering because none of them have all these three properties.

We provide more details of the three properties in the following three subsections. The key contender to the proposed framework is the coreset-based $k$-means based Framework ${\dddot{\mathbbm{B}}}$ because it is the only one which has property (a) (as mentioned in Section \ref{sec-first-approach}). We focus our comparison with it in the rest of this paper.

\subsection{Distributed-Centralized Clustering Equivalence}
\label{sec-equivalence}
The proposed two frameworks have the distributed-centralized clustering equivalence property when they use the same subset to perform the clustering in step~2. This yields, $\forall i,\  \mathbb{C}_i = \bigcup_\ell \mathbb{C}_i^\ell$.

This outcome can be achieved when the sampled subsets $\mathcal{B}$ and $\bigcup_l\mathcal{B}_l$, from ${\mathbbm{A}}$ and $\dddot{\mathbbm{A}}$, respectively, have the same points. 

This outcome remains, irrespective of the data sizes and cluster
distributions on different sites in a network. 

Among all the existing methods of distributed clustering, only the coreset-based $k$-means distributed clustering has property (a). 

Other existing methods  approximate the clustering outcome of their centralized counterpart only, with or without clustering performance guarantee. They do not have property (a) because the clustering outcomes of the distributed clustering are affected by the dataset size on each site and the number of sites, as well as the clustering distributions on these sites. 

In addition, these methods often require that the given dataset is evenly distributed over all sites. The  of these methods degrades when the dataset is unevenly distributed
In addition, these methods often require that the given dataset is evenly distributed over all sites. The clustering outcomes of these methods degrade when the dataset is unevenly distributed.
% In summary, 

\subsection{Communication Cost and Time Complexity}

In a distributed clustering framework, the overall runtime consists of the total communication time between the sites and the coordinator, and the maximum runtime at any site.

\textbf{Communication cost.} Ideally, a distributed clustering framework shall have a small constant communication cost.

For ${\dddot{\mathbbm{A}}}$, the entire process has the same constant communication cost, irrespective of the actual clustering $\mathbbm{f}$ used in step 2. For ${\dddot{\mathbbm{B}}}$, only coreset-based $k$-means \cite{NIPS2013_Balcan,NEURIPS2018_Li,DR-kmeans-2015,ipca14} can guarantee constant communication cost. A comparison between these two frameworks is shown the last column in Table \ref{tab-time-complexity}.

In both frameworks, they require $r$ sites to pass a total of $s$ data points to the coordinator at the end of step 1; and the coordinator to pass $k$ (feature mapped) centers to each site at the end of step 2. The total communication cost is $s+kr$. In addition, the coordinator in ${\dddot{\mathbbm{A}}}$ needs to deliver a total of $\chi r$ kernel mean maps to $r$ sites\footnote{When Isolation kernel (IK) is used, the cost of building and using its feature map $\chi = \psi t$, where $\psi$ is the number of partitions in a partitioning, and $t$ is the number of partitionings used to derive IK (see \cite{ting2018IsolationKernel} for details).}. Since the feature mapping needs to be performed first for ${\dddot{\mathbbm{B}}}$-$\kappa k$m before the coreset is calculated, $s$ additional communication costs are required.

Note that all the methods shown in Table \ref{tab-time-complexity} have constant communication cost.

\textbf{Time complexity.} A distributed clustering shall have linear or near-linear time complexity in order to deal with big data.

Existing methods in Framework ${\dddot{\mathbbm{B}}}$ use a clustering algorithm (sometime with quadratic time complexity) in step 1 to find a suitable subset. Table \ref{tab-time-complexity} shows the time complexity $(log(n))^{poly(k/\epsilon)}$ of the coreset method \cite{coreset-time} used in step 1 which is the most costly step in ${\dddot{\mathbbm{B}}}$.  The proposed Framework ${\dddot{\mathbbm{A}}}$ spends the minimum amount of time in step 1 to get a random subset, unlike ${\dddot{\mathbbm{B}}}$.

Step 2 performs clustering on the combined set of subsets obtained from all sites, and its data size is much smaller than the total data size from all sites, i.e., $s \ll n$. The time complexity depends on the clustering algorithm used and the data size $s$. This is the same for both ${\dddot{\mathbbm{A}}}$ and ${\dddot{\mathbbm{B}}}$.  %Yet, the time cost of ${\dddot{\mathbbm{A}}}$ in step 3 could be larger than that in step 2 because $s$ is fixed which is often much smaller than $n$.

In practice, step 3 in both ${\dddot{\mathbbm{A}}}$ and ${\dddot{\mathbbm{B}}}$  costs $nk/r$ (assuming data size is evenly distributed across all sites). If the data size is unevenly distributed, the cost is the time on the site having the largest data size. This is the same for any of the ${\dddot{\mathbbm{A}}}$ and ${\dddot{\mathbbm{B}}}$  frameworks.

Interestingly, the proposed ${\dddot{\mathbbm{A}}}$ satisfies property (b) independent of the following:
\begin{itemize}
    \item the number of sites $r>1$,
    \item whether the data size is evenly or unevenly distributed over all sites,
    \item cluster distributions on the $r$ sites.
\end{itemize}

On the contrary, the step 1  runtime of ${\dddot{\mathbbm{B}}}$ is influenced by the number of distributed sites and the distribution of data size among the sites. As a result, the overall runtime can exceed the runtime of its centralized counterpart. Therefore, ${\dddot{\mathbbm{B}}}$ does not have property (b). We evaluate this in the experimental section.

In summary, the determinant in meeting property (b) is the time complexity of the process in step 1, not step 2 or 3. Because  $\dddot{\mathbbm{B}}$ has its most costly component in step 1, it usually does not satisfy property (b). The proposed Framework $\dddot{\mathbbm{A}}$ has property (b) for the opposite reason---having the least costly component in step~1.

\begin{table}[t]
\caption{Worst case time complexities and communication costs in $\dddot{\mathbbm{A}}$ and $\dddot{\mathbbm{B}}$. The clustering algorithms ($\mathbbm{f}$ in step 2) are (i) $\kappa$BCC: our proposed Kernel Bounded Cluster Cores (see Section \ref{sec-kappa}); (ii) $\kappa k$m: kernel $k$-means \cite{complexity_kkm} ; (iii) $k$m: $k$-means \cite{k-means-SCG2006}. $s = |\mathcal{B}|$ is the sample size used to perform clustering in step 2. $n= |D|$ is the total data size from all sites. $r$ is the total number of sites (excluding the coordinator). $k$ is the number of final clusters. $\epsilon$ is the approximate error in coreset \cite{coreset-time}. $\beta$ is the number of iterations required in kernel $k$-means. $\chi$ is the cost of building and using the feature map of kernel.}
\label{tab-time-complexity}
\scalebox{1.0}{
\centering
\begin{tabular}{l|ccc|c|l}
\hline
      & \multicolumn{4}{c|}{\hspace*{1.5cm}Time complexity}   & \multirow{2}[1]{*}{Comm. cost} \\ \cline{2-5}
   & \multicolumn{3}{c|}{Distributed} & \multicolumn{1}{c|}{Centralized} & \\ \cline{2-4}
                                          & Step 1                                                  & Step 2    & Step 3 &                                               &   \\ \hline
 $\dddot{\mathbbm{A}}$-$\kappa$BCC        & $s$                                                     & $s^2$                    & $nk$ &  $n+s^2$                                        & $s+(k+\chi)r$ \\ 
%$k$m-s    & polynomial time   & $s+kr$ \\ \hline
  ${\dddot{\mathbbm{A}}}$-$\kappa k$m     & $s$                                                     & $ s^2\beta$   & $nk$ & $n^2\beta$                & $s+(k+\chi)r$ \\ \hline
  ${\dddot{\mathbbm{B}}}$-$\kappa k$m     & $(log(n))^{poly(k/\epsilon)}$                           & $ s^2\beta$              & $nk$ & $n^2\beta$                                      & $2s+(k+\chi)r$ \\ 
 ${\dddot{\mathbbm{B}}}$-$k$m             & $(log(n))^{poly(k/\epsilon)}$                           & $ 2^{\Omega(\sqrt{s})}$  & $nk$ & $2^{\Omega(\sqrt{n})}$    & $s+kr$ \\ \hline
\end{tabular}
}
\end{table}

\subsection{Clustering Performance Guarantee}
\label{sec-property-c}
The proposed Frameworks ${\mathbbm{A}}$ and $\dddot{\mathbbm{A}}$ admit clusters of arbitrary shapes, sizes and densities because step 3 is a distribution-based point assignment step. We show here that the frameworks find a good approximates of the ground truth clusters $T_i$ if the pdf of initial clusters $P_{G_i}$ approximates the pdf of the ground truth clusters $P_{T_i}$, i.e., $P_{G_i}\approx P_{T_i}$.

\begin{prop}
Assume that every point $\mathbf{x}$ in the given dataset $D$ belongs to only one of $k$ ground truth clusters $\mathcal{T}_i$, and they are non-overlapping clusters, i.e., $\mathcal{T}_i \cap \mathcal{T}_j = \emptyset\ \forall i \ne j$.
%are separable in the the feature space of distributional kernel $K$ according to the Voronoi diagram as stated Definition \ref{def_clusters}.

The clusters $C_i$ produced from $D$ using the initial clusters $\mathcal{G}_i$ via the distribution-based point assignment:
\[
%\mbox{For } i =1,\dots,k: 
C_i=\left\{\mathbf{x}\in D\ |\ \mathop{\argmax}\limits_{j \in [1,k]} \mathcal{K}(\delta (\mathbf{x}), \mathcal{P}_{\mathcal{G}_j})=i \right\}
\]
%i.e., \[\mbox{For } j =1,\dots,k: \mathcal{T}_j=\{\mathbf{x}\in D\ |\ \mathop{\argmax}\limits_{i \in [1,k]} \mathcal{K}(\delta (\mathbf{x}), \mathcal{P}_\mathcal{T}_i)=j \},   \]
$\mbox{ has } \mathcal{P}_{\mathcal{C}_i} \approx \mathcal{P}_{\mathcal{T}_i} \mbox{ if } \mathcal{P}_{\mathcal{G}_i} \approx \mathcal{P}_{\mathcal{T}_i}\ \forall i$.
\end{prop}

The clustering outcome requires that the pdf of the discovered cluster $C_i$ approximates the pdf of the ground truth cluster $\mathcal{T}_i$, i.e., $P_{C_i}\approx P_{\mathcal{T}_i}$. This result follows directly from the distribution-based point assignment using the distributional kernel $\mathcal{K}$. It assigns each point to the most similar distribution of the initial cluster $G_i$---yielding the pdf of the discovered cluster $C_i$ which approximates the pdf of the initial cluster $G_i$: $P_{C_i}\approx P_{G_i}$. To get this last result, the pre-requisite is that the distribution of the initial cluster $G_i$ approximates  the pdf of the ground truth cluster $\mathcal{T}_i$: $\mathcal{P}_{\mathcal{G}_i} \approx \mathcal{P}_{\mathcal{T}_i}$.

\begin{figure}[b]
\setlength\tabcolsep{1pt}
\scalebox{1}{
% \footnotesize
    \centering
    \begin{tabular}{l|cc|cc}
         \hline
 & \multicolumn{2}{c|}{Jain} & \multicolumn{2}{c}{Complex9} \\ \midrule
        
Step~2  & \multicolumn{2}{c|}{$\mathcal{P}_{\mathcal{G}_i} \approx \mathcal{P}_{\mathcal{T}_i}$} & \multicolumn{2}{c}{$\mathcal{P}_{\mathcal{G}_i} \approx \mathcal{P}_{\mathcal{T}_i}$} \\
&   \multicolumn{2}{c|}{\begin{minipage}[b]{0.225\columnwidth}
             \centering
             \raisebox{-.5\height}{\includegraphics[width=\linewidth]{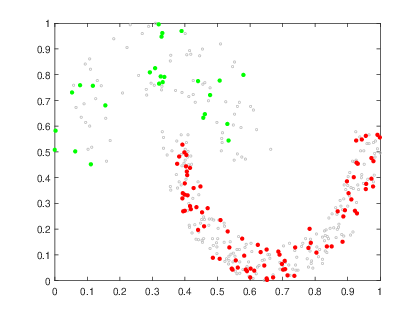}}
         \end{minipage}}
         & \multicolumn{2}{c}{\begin{minipage}[b]{0.225\columnwidth}
             \centering
             \raisebox{-.5\height}{\includegraphics[width=\linewidth]{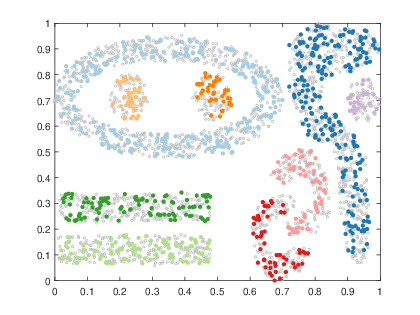}}
         \end{minipage}}
         \\\midrule
      Step~3  & $\mathcal{K}(\delta(\mathbf{x}),\mathcal{P}_{\mathcal{G}_i})$  & $\Vert \mathbf{x} - \bar{\mathbf{x}}_{\mathcal{G}_i} \Vert$ & $\mathcal{K}(\delta(\mathbf{x}),\mathcal{P}_{\mathcal{G}_i})$ & $\Vert \mathbf{x} - \bar{\mathbf{x}}_{\mathcal{G}_i} \Vert$  \\    

 &   \begin{minipage}[b]{0.225\columnwidth}
             \centering
             \raisebox{-.5\height}{\includegraphics[width=\linewidth]{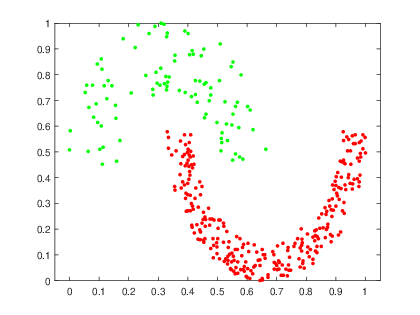}}
         \end{minipage} 
        &   \begin{minipage}[b]{0.225\columnwidth}
             \centering
             \raisebox{-.5\height}{\includegraphics[width=\linewidth]{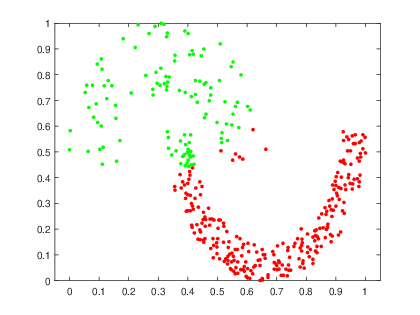}}
         \end{minipage} 
        &   \begin{minipage}[b]{0.225\columnwidth}
             \centering
             \raisebox{-.5\height}{\includegraphics[width=\linewidth]{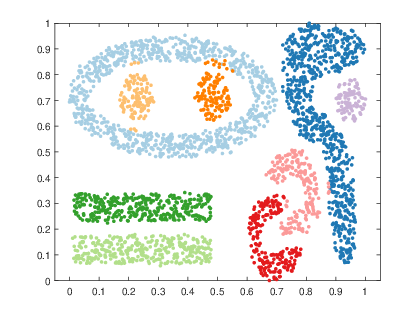}}
         \end{minipage} 
        &   \begin{minipage}[b]{0.225\columnwidth}
             \centering
             \raisebox{-.5\height}{\includegraphics[width=\linewidth]{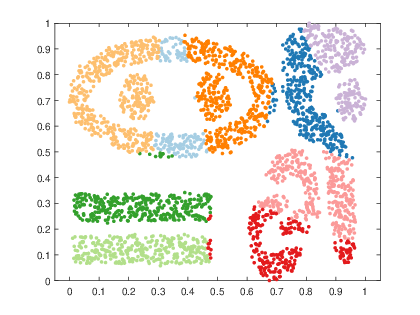}}
         \end{minipage} 
         \\ 
\multicolumn{1}{r|}{NMI \ \ \ }      & 1.00   & 0.55  &  0.98  & 0.71    
         \\ \hline
    \end{tabular}
    }
    \caption{The impact of using initial clusters $\mathcal{G}_i$ having $\mathcal{P}_{\mathcal{G}_i} \approx \mathcal{P}_{\mathcal{T}_i}$ in step 2; and compare distribution-based point assignment $\mathcal{K}(\delta(\mathbf{x}),\mathcal{P}_{\mathcal{G}_i})$  with center-based point assignment $\Vert \mathbf{x} - \bar{\mathbf{x}}_{\mathcal{G}_i} \Vert$  in step 3. Either the proposed Framework ${\mathbbm{A}}$ or ${\dddot{\mathbbm{A}}}$ produces the same clustering outcomes. NMI: normalized mutual information  \cite{NMI2010}
    }
  \label{fig:step3-change}
\end{figure}

Figure \ref{fig:step3-change} provides examples of clustering outcomes when a random sample of $\mathcal{P}_{\mathcal{G}_i} \approx \mathcal{P}_{\mathcal{T}_i}$ is used in step 2 in the proposed framework (either ${\mathbbm{A}}$ or ${\dddot{\mathbbm{A}}}$ produces exactly the same outcome). On both datasets, the distribution-based point assignment is able to produce a perfect or near-perfect clustering outcome.
%with the Isolation kernel function. 

Figure \ref{fig:step3-change} also shows that center-based point assignment (as used in $k$-means in existing Framework ${\dddot{\mathbbm{B}}}$) fails to find the appropriate clusters even though the same sample $\mathcal{G}_i$ is used in step 2. This is because the center-based point assignment does not consider the cluster distributions.

Note that the distributed-based point assignment based on a distributional kernel has its theoretical underpinning on kernel mean embedding of distribution  \cite{KernelMeanEmbedding2017}.

While it is not always possible to ensure that $\mathcal{P}_{\mathcal{G}_i} \approx \mathcal{P}_{\mathcal{T}_i}$, $\mathcal{K}$DC produces a good approximation to the ground truth cluster as long as $\mathcal{G}_i$ is a good quality initial cluster. This requires a clustering algorithm that can find clusters such that $\mathcal{P}_{\mathcal{G}_i} \approx \mathcal{P}_{\mathcal{T}_i}$.

An example of a good quality initial cluster consists of high similarity points in a subset of $\mathcal{T}_i$. A way to obtain this kind of initial cluster is presented in the next section.

\section{Proposed Kernel-Bounded Cluster Cores}
\label{sec-kappa}

Here we propose to use a new clustering algorithm called Kernel Bounded Cluster Cores ($\kappa$BCC) as $\mathbbm{f}$ in $\mathbbm{A}$ and $\dddot{\mathbbm{A}}$. 

\begin{definition}
Given a dataset $D$, $\kappa$BCC produces the $k$ largest  $\kappa_\tau$-cluster cores $\mathcal{G}$, each encapsulates the `core' of a cluster, defined based on a kernel $\kappa$ with a threshold $\tau$, as follows:
\[
\begin{aligned}
\mathcal{G} =\{\mathbf{x},\mathbf{y}\in D\ |\ \mbox{there exists a chain: } \mathbf{z}_1,\mathbf{z}_2,\cdots,\mathbf{z}_j,\ 
 \mbox{such that } \mathbf{z}_1= \mathbf{x}, \mathbf{z}_j=\mathbf{y}, \forall i\ \kappa(\mathbf{z}_i, \mathbf{z}_{i+1}) > \tau \}
\end{aligned}
\]
\end{definition}

Intuitively, $\mathcal{K}$DC first samples a subset from the original data, the distribution of which is very close to the distribution of the original data. Then $\kappa$BCC is used to find $\mathcal{G}$, which contains chains of core points with very high similarity. These points are often in the high-density regions of each cluster ($P_\mathcal{G}$ in Figure \ref{fig:illus}), and then the points are assigned according to the similarity with the distribution of these high-density areas.

Note that $\kappa$BCC is different from (kernel) $k$-means in two ways. First, $\kappa$BCC is not strictly a clustering algorithm because its final clustering outcome does not cover all points in the given dataset, i.e., $\bigcup_i \mathcal{G}_i \subset D$. This is because the process potentially produces the number of clusters more than $k$, and only the points having similarity higher than $\tau$ are included in each $\mathcal{G}_i$. Second, more importantly, the $\kappa_\tau$-clusters have arbitrary shapes, sizes and densities, adhering to the data distribution in the dataset. This enables the final clustering outcome, after the subsequent point assignment in step 3, to discover clusters having the same distributional characteristics found in the given dataset. 

\subsection{Determinants of $\mathcal{K}$DC Property (c)}

\begin{figure}[b]
\setlength\tabcolsep{2pt}
\scalebox{1}{
% \footnotesize
    \centering
    \begin{tabular}{l|c|c|c}
         \hline

    Jain  & ~  & ~ & ~ \\   
Step~2   &$k$-means & DP &  $\kappa$BCC \\ 
   &  \begin{minipage}[b]{0.285\columnwidth}
             \centering
             \raisebox{-.5\height}{\includegraphics[width=\linewidth]{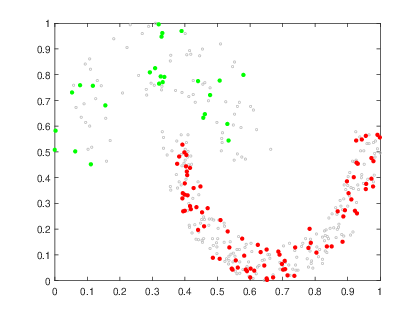}}
         \end{minipage}
        &   \begin{minipage}[b]{0.285\columnwidth}
             \centering
             \raisebox{-.5\height}{\includegraphics[width=\linewidth]{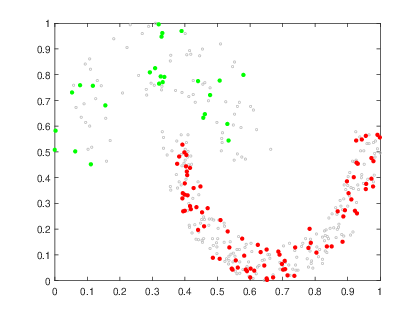}}
         \end{minipage}
        &   \begin{minipage}[b]{0.285\columnwidth}
             \centering
             \raisebox{-.5\height}{\includegraphics[width=\linewidth]{graphs/jain_KBC_kernel_step2_nmi_1.png}}
         \end{minipage}
         \\ \cline{2-4}
Step~3  & \multicolumn{3}{c}{-------------- $\mathcal{K}(\delta(\mathbf{x}),\mathcal{P}_{\mathcal{G}_i})$ ---------------} \\ 
 &  \begin{minipage}[b]{0.285\columnwidth}
             \centering
             \raisebox{-.5\height}{\includegraphics[width=\linewidth]{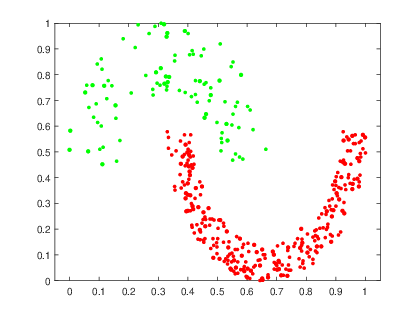}}
         \end{minipage}
        &   \begin{minipage}[b]{0.285\columnwidth}
             \centering
             \raisebox{-.5\height}{\includegraphics[width=\linewidth]{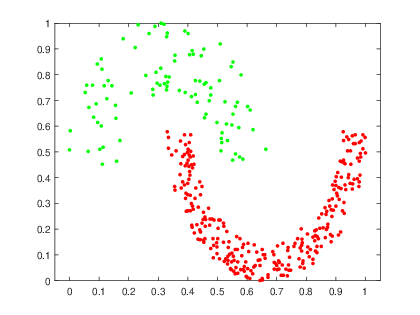}}
         \end{minipage}
        &   \begin{minipage}[b]{0.285\columnwidth}
             \centering
             \raisebox{-.5\height}{\includegraphics[width=\linewidth]{graphs/jain_KBC_kernel_step3_nmi_1.png}}
         \end{minipage}  
         \\ 
\multicolumn{1}{r|}{NMI\ \ \ }     & 1.00   & 1.00  & 1.00 
         \\ \hline
    Complex9  & ~  & ~ & ~  \\   
Step~2  & \multicolumn{1}{c|}{$k$-means} & \multicolumn{1}{c|}{ DP} & \multicolumn{1}{c}{ $\kappa$BCC} \\   
   &  \begin{minipage}[b]{0.285\columnwidth}
             \centering
             \raisebox{-.5\height}{\includegraphics[width=\linewidth]{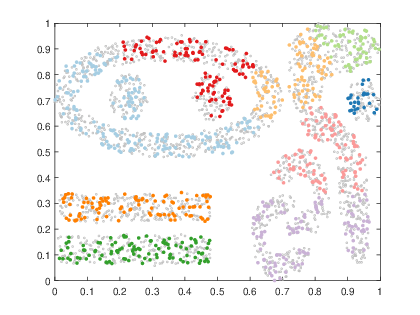}}
         \end{minipage}
        &   \begin{minipage}[b]{0.285\columnwidth}
             \centering
             \raisebox{-.5\height}{\includegraphics[width=\linewidth]{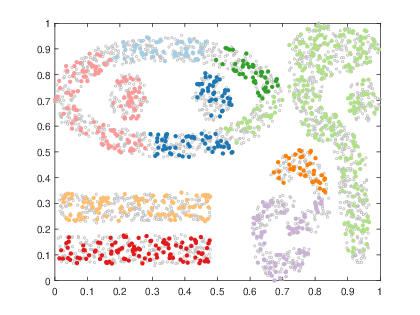}}
         \end{minipage}
        &   \begin{minipage}[b]{0.285\columnwidth}
             \centering
             \raisebox{-.5\height}{\includegraphics[width=\linewidth]{graphs/com9_KBC_kernel_step2_nmi_098.png}}
         \end{minipage}
         \\ \cline{2-4}
Step~3  & \multicolumn{3}{c}{-------------- $\mathcal{K}(\delta(\mathbf{x}),\mathcal{P}_{\mathcal{G}_i})$ ---------------} \\  
 &  \begin{minipage}[b]{0.285\columnwidth}
             \centering
             \raisebox{-.5\height}{\includegraphics[width=\linewidth]{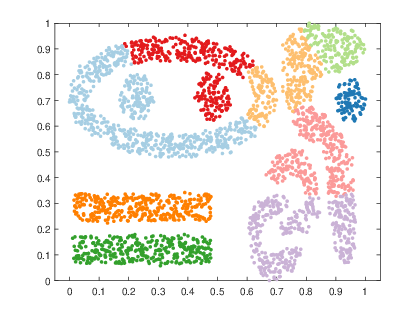}}
         \end{minipage}
        &   \begin{minipage}[b]{0.285\columnwidth}
             \centering
             \raisebox{-.5\height}{\includegraphics[width=\linewidth]{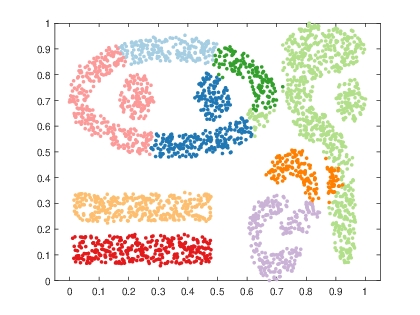}}
         \end{minipage}
        &   \begin{minipage}[b]{0.285\columnwidth}
             \centering
             \raisebox{-.5\height}{\includegraphics[width=\linewidth]{graphs/com9_KBC_kernel_step3_nmi_098.png}}
         \end{minipage} 
         \\ 
\multicolumn{1}{r|}{NMI\ \ \ }     & 0.72  & 0.77  & 0.98    \\ \hline
    \end{tabular}
    }
    \caption{The clustering outcomes of $\mathcal{K}$DC or $\dddot{\mathbbm{A}}$ in steps 2 \& 3 (and also in terms of NMI) using $k$-means, DP and $\kappa$BCC in step 2 on two datasets: Jain (top) and Complex9 (bottom). The final clustering outcomes are shown in the `Step 3' row ($s=0.3n$ is used).}
  \label{fig:step2-change}
\end{figure}

Figure \ref{fig:step2-change} (last column) shows the example clustering outcomes as a result of using $\kappa$BCC in step 2 in either of the two proposed Frameworks ${\mathbbm{A}}$ and ${\dddot{\mathbbm{A}}}$.

As step 2 in the Frameworks admits any clustering algorithm, Figure \ref{fig:step2-change} also shows the example clustering outcomes of using either $k$-means or DP \cite{DP2014} in step 2. While they enable perfect clustering outcomes on the simple Jain dataset, they fail to do so on the Complex9 dataset. 

The examples in Figures \ref{fig:step3-change} and \ref{fig:step2-change} show that there are two determinants in discovering clusters of arbitrary shapes, varying data sizes and densities in order to have  property (c) of $\mathcal{K}$DC:
\begin{itemize}
    \item Initial clusters $\mathcal{G}_i$ that represent the core of the true clusters, and 
    \item The point assignment procedure that enables  clusters of arbitrary shapes, sizes and densities to be found:
\[C_{j}=\left\{\mathbf{x}\in D\ |\ \mathop{\argmin}\limits_{i \in [1,k]} \mathcal{K}(\delta(\mathbf{x}),\mathcal{P}_{\mathcal{G}_i}) =j \right\} \]
\end{itemize}
They are in steps 2 and 3 of the two proposed Frameworks.\\
Note that replacing step 3 with the typical center-based point assignment of $k$-means disables this capability, regardless of how good the initial clusters are, as shown in Figure \ref{fig:step3-change}.

In a nutshell, step 3 is the enabling determinant in satisfying property (c), and step 2 is the supporting determinant. Getting a sufficient sample size in step 1 is a necessary factor too.

\subsection{A Comparison of $\kappa$BCC and SOTA Algorithm}

Recall that coreset-based $k$-means is the SOTA distributed clustering method of $\dddot{\mathbbm{B}}$.

$\kappa$BCC has two key advantages in either $\mathbbm{A}$ or $\dddot{\mathbbm{A}}$ over the coreset-based $\dddot{\mathbbm{B}}$:

\begin{enumerate}
    \item The subset $\mathcal{B}$ in step 1 of either Framework is simply a random subset of $S$ (or $\bigcup_\ell D_\ell$), rather than a coreset. The latter requires a computationally expensive process and it is tightly coupled with the clustering algorithm used. A random subset needs none of those.
    
    \item The initial $k$ representatives of $\kappa$BCC are determined via similarity-based clustering. At the end of step 2, the summarized centers of these $k$ high-similarity clusters (via kernel mean embedding \cite{KernelMeanEmbedding2017}) are broadcast to each site to assign points to the most similar distribution in one iteration independently. As a result, the clusters are not restricted to globular shape, equal size and equal density---the only type of clusters that can be produced by $k$-means in Framework ${\dddot{\mathbbm{B}}}$. 
    
    Put in another way, a coreset is required only if (kernel) $k$-means is used in either $\mathbbm{A}$ or $\dddot{\mathbbm{A}}$ because it stabilizes the clustering outcome. A random subset, instead of a coreset, will produce a wildly different clustering outcome compared with that produced from a different random subset. Yet, $\kappa$BCC produces stable $\kappa_\tau$-clusters provided the random subset is a sufficient representative sample of the data distribution.
\end{enumerate}

It is possible to improve $k$-means by using kernel $k$-means (shown in the experiment section). But the key issues are that (i) kernel $k$-means still performs poorer than $\kappa$BCC when evaluated on an entire dataset in the centralized clustering setting (see Section \ref{sec-centralized-comparison} for details); and (ii) instead of spending a lot of time computing the coreset, to make it scalable, kernel $k$-means inevitably uses an approximation approach (e.g., employ a small sample set). This approximation further degrades their clustering performance. In other words, the approach is to trade off effectiveness with efficiency. $\kappa$BCC has no such issue. Though $\kappa$BCC also uses a small sample,  coupled with step 3 as a centralized clustering shown in Framework ${\mathbbm{A}}$, it is able to find clusters of arbitrary shapes, sizes and densities, without compromising the clustering quality. 
%produces a better outcome than large samples (see the discussion in the IK-t-SNE paper\cite{IK-t-SNE}.)

\section{Experimental Designs and Settings}

The experiments are designed with the following aims: 

\begin{enumerate}
    \item Compare the relative performance of Frameworks $\dddot{\mathbbm{A}}$ and $\dddot{\mathbbm{B}}$ in terms of NMI (normalized mutual information  \cite{NMI2010}), AMI (Adjusted Mutual Information \cite{AMI}), F1 \cite{christen2023review}, ARI (Adjusted Rand Index \cite{steinley2004properties}) and runtime\footnote{Detailed results of AMI, F1 and ARI are provided in the supplementary materials and \url{https://anonymous.4open.science/r/KDC-kbcc/}.}.
    \item Verify  property (b) of Framework $\dddot{\mathbbm{A}}$ or $\mathcal{K}$DC.
    \item Investigate $\dddot{\mathbbm{A}}$ as a generic framework that enables any quadratic time clustering algorithms to deal with large datasets.
    \item Examine the relative performance of four methods of centralized clustering.
\end{enumerate}

Specifically, the proposed $\kappa$BCC\footnote{The source code, the data characteristics of the datasets we used and parameter search ranges of all methods in the experiments are provided at \url{https://anonymous.4open.science/r/KDC-kbcc/}.} and kernel $k$-means are used as $\mathbbm{f}$ in $\dddot{\mathbbm{A}}$. In $\dddot{\mathbbm{B}}$, coreset-based $k$-means \cite{NIPS2013_Balcan} is chosen as the representative algorithm for two reasons. First, coreset-based $k$-means is the only algorithm in $\dddot{\mathbbm{B}}$ that satisfies the property (a) with a theoretical guarantee. Second, coreset-based $k$-means is the only algorithm in $\dddot{\mathbbm{B}}$ with a deterministic constant communication cost, which is the same as $\dddot{\mathbbm{A}}$. In addition, we also implemented a kernel version of coreset-based $k$-means for comparison.

The empirical evaluation is conducted using seven datasets from \url{https://archive.ics.uci.edu/}.

\textbf{Experimental details}. 
For each dataset, we first simulate a communication network connecting $r$ local sites as conducted by previous work  \cite{NIPS2013_Balcan}, and
then partition the dataset into local data subsets. If not explicitly stated, the data sizes at all sites are evenly distributed. In the experiments, $r$=20 sites and the subset data size at the end of step 1 is $s$=$min(n,10000)$. 
All data are normalized to the range [0, 1] in the pre-processing. 

Unless otherwise specified, the kernel used in a kernel-based clustering algorithm is Isolation kernel \cite{ting2018IsolationKernel}.

The experiments were executed on a Linux CPU machine: AMD 128-core CPU with each core running at 2 GHz and  1T GB RAM. For each method, we report the average result of five trials.

The results of the four experiments on distributed clustering, which correspond to the above four aims, are reported in the next section, and the evaluation results of four methods of centralized clustering (the fifth aim) are presented in Section \ref{sec-centralized-comparison}.

\section{Distributed Clustering Evaluation}
\subsection{Performances of Frameworks ${\dddot{\mathbbm{A}}}$ and ${\dddot{\mathbbm{B}}}$}
\label{sec-exp1}

\begin{figure}[t]
    \centering
    \includegraphics[width=0.6\textwidth]{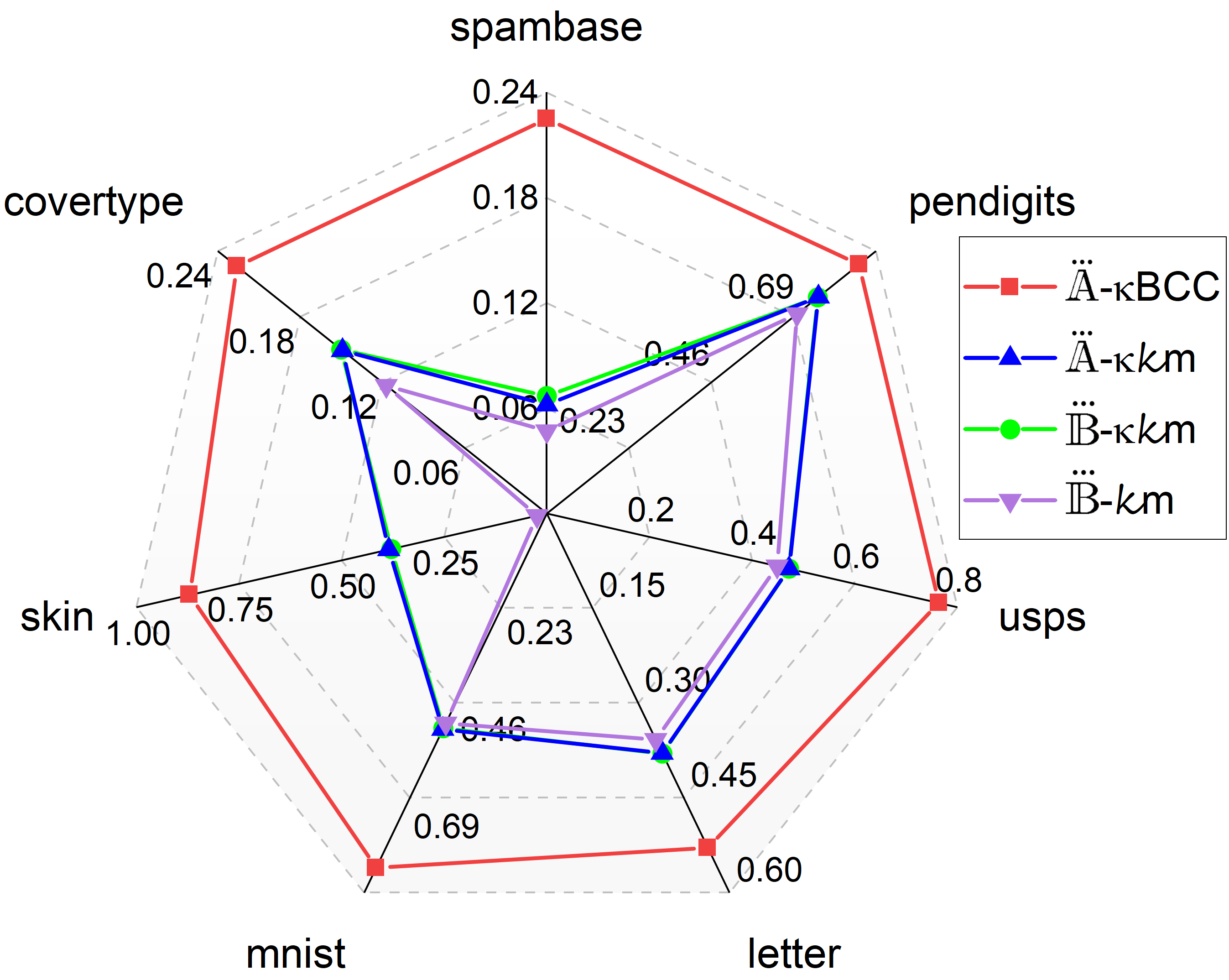}
    \caption{The comparison of ${\dddot{\mathbbm{A}}}$-$\kappa$BCC, ${\dddot{\mathbbm{A}}}$-$\kappa k$m (kernel $k$-means), coreset $k$-means in the existing framework and its kernel version (${\dddot{\mathbbm{B}}}$-$k$m and ${\dddot{\mathbbm{B}}}$-$\kappa k$m) in terms of NMI. 
    }
    \label{fig:main_ik}
\end{figure}

\textbf{Clustering results} The clustering results of the comparison are shown in Figure \ref{fig:main_ik}. 

The results show that  ${\dddot{\mathbbm{A}}}$-$\kappa$BCC achieves the best results on all seven datasets,. On average, compared with the existing state-of-the-art algorithm ${\dddot{\mathbbm{B}}}$-$k$m, our method achieves $685\%$ improvement in terms of NMI. Compared with ${\dddot{\mathbbm{B}}}$-$\kappa k$m and ${\dddot{\mathbbm{A}}}$-$\kappa k$m, ${\dddot{\mathbbm{A}}}$-$\kappa$BCC also achieves $113\%$ and $116\%$ improvements, respectively. Especially, on the skin dataset, ${\dddot{\mathbbm{A}}}$-$\kappa$BCC achieves $3787\%$, $253\%$ and $244\%$ improvements over ${\dddot{\mathbbm{B}}}$-$k$m, ${\dddot{\mathbbm{B}}}$-$\kappa k$m and ${\dddot{\mathbbm{A}}}$-$\kappa k$m, respectively. 

Note that ${\dddot{\mathbbm{A}}}$-$\kappa k$m and ${\dddot{\mathbbm{B}}}$-$\kappa k$m have comparable clustering results. The key difference between them is the use of coreset in ${\dddot{\mathbbm{B}}}$ and a random subset in ${\dddot{\mathbbm{A}}}$.  Although some existing studies \cite{NIPS2013_Balcan,NEURIPS2018_Li,ipca14} have focused on the importance of coreset, our results show that there is no significant difference in the clustering outcomes from a coreset or random subset.

In a nutshell, $\mathcal{K}$DC outperforms the $k$-means based framework. $\kappa$BCC is a better clustering algorithm than kernel $k$-means under $\mathcal{K}$DC because $\kappa$BCC does not have the fundamental limitations of (kernel) $k$-means. Spending a lot of time finding a coreset has no real payoff.
This result verifies that the focus of a distributed clustering framework shall not be in step 1.

\begin{figure}[b]
    \centering
    \includegraphics[width=0.64\textwidth]{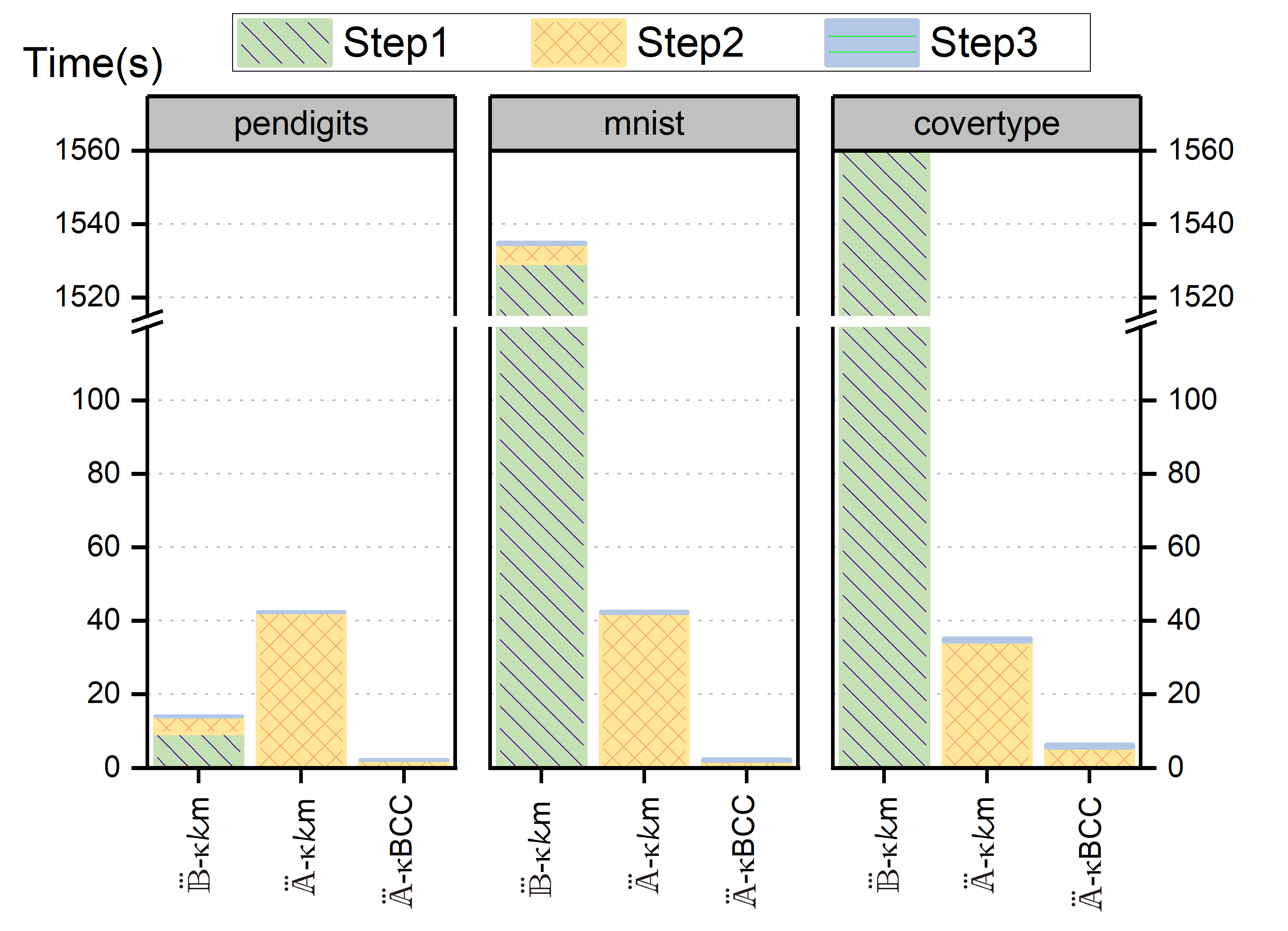}
    \caption{The comparison of runtimes of ${\dddot{\mathbbm{B}}}$-$\kappa k$m, ${\dddot{\mathbbm{A}}}$-$\kappa k$m and ${\dddot{\mathbbm{A}}}$-$\kappa$BCC on the pendigits, mnist and covertype datasets.}
    \label{fig:main_time}
\end{figure}

\textbf{Runtime}. Figure \ref{fig:main_time} shows the runtime comparison of the three clustering algorithms  ${\dddot{\mathbbm{B}}}$-$\kappa k$m, ${\dddot{\mathbbm{A}}}$-$\kappa k$m and ${\dddot{\mathbbm{A}}}$-$\kappa$BCC.

%A more comprehensive analysis of the runtime results is provided below. 
The coreset-based ${\dddot{\mathbbm{B}}}$-$\kappa k$m took the longest time because it spent a lot of time in step 1 to calculate the coreset, and could not complete in two days on the large covertype dataset. Note that the coreset computational time took much longer than the time to run kernel $k$-means on large datasets, i.e., the cure is worse than the disease---considering that distributed clustering is meant to reduce the runtime of centralized clustering. 
% \textcolor{red}{Is this correct? Fig 3 is the total runtime cost or the maximum runtime cost of any site?}

In contrast, ${\dddot{\mathbbm{A}}}$-$\kappa k$m took a minimum amount of time in step 1 while it has the longest runtime in step 2 because the optimization took longer to converge for a random subset than a coreset. However, the total runtime of ${\dddot{\mathbbm{A}}}$-$\kappa k$m is still significantly shorter than that of ${\dddot{\mathbbm{B}}}$-$\kappa k$m on the two large datasets (mnist and covertype).

${\dddot{\mathbbm{A}}}$-$\kappa$BCC has the shortest runtimes in both steps 1 and 2. Its total runtime is one order of magnitude faster than ${\dddot{\mathbbm{A}}}$-$\kappa k$m, and at least four orders of magnitude faster than ${\dddot{\mathbbm{B}}}$-$\kappa k$m on the large covertype dataset.

Note that because the size of subset $\mathcal{B}$ is fixed, the runtime is almost the same for each clustering algorithm (in step 2) in either ${\dddot{\mathbbm{A}}}$ or ${\dddot{\mathbbm{B}}}$  on any dataset.

The above results can be summarized in three points. First, the coreset-based method is  not a practical method because a coreset and a random subset produce comparable NMI clustering outcomes, despite spending a lot of time calculating a coreset. Second, $\mathcal{K}$DC or Framework ${\dddot{\mathbbm{A}}}$  takes the least amount of time in step 1;  the time spent in step 2 is independent of the total data size if the subset data size $s$ is fixed; and the time complexity in step 3 is linear to the total data size. Third, $\kappa$BCC is a clustering algorithm that uses no optimization, making its runtime (in step 2 of a framework) shorter than kernel $k$-means.

\subsection{$\mathcal{K}$DC Satisfies Property (b)}
\label{sec-experiment-property(b)}

Table \ref{qdcs_b} shows the comparison of the runtimes of the distributed $\dddot{\mathbbm{A}}$ and the centralized ${\mathbbm{A}}$ frameworks (both using $\kappa$BCC) on four datasets of different sizes.

\begin{table}[h]
\centering
\caption{Runtime comparison (in seconds) of the distributed $\dddot{\mathbbm{A}}$ framework and the centralized ${\mathbbm{A}}$ framework}
\label{qdcs_b}
\begin{tabular}{lrr|r|r}
\hline
Dataset & \#$n$ & \#$d$  & Distributed $\dddot{\mathbbm{A}}$ & Centralized ${\mathbbm{A}}$    \\ \hline
pendigits  & 10,992 & 16  & 2.0        & 2.1       \\ 
mnist  &  70,000 & 784 & 2.6        & 12.0      \\ 
covertype &  581,012 & 54 & 5.8        & 28.4       \\ 
mnist5m & 5,000,000 & 784 & 36\hspace{2.5mm} & 480\hspace*{2.5mm}\\ \hline
\end{tabular}
\end{table}

The results show that the runtime of the distributed $\dddot{\mathbbm{A}}$ is lower than that of the centralized ${\mathbbm{A}}$. This is because in the distributed mode, the point assignment (step 3) is computed on multiple machines in parallel, thus reducing the overall time.

\begin{figure}[t]
    \centering
    \includegraphics[width=0.5\textwidth]{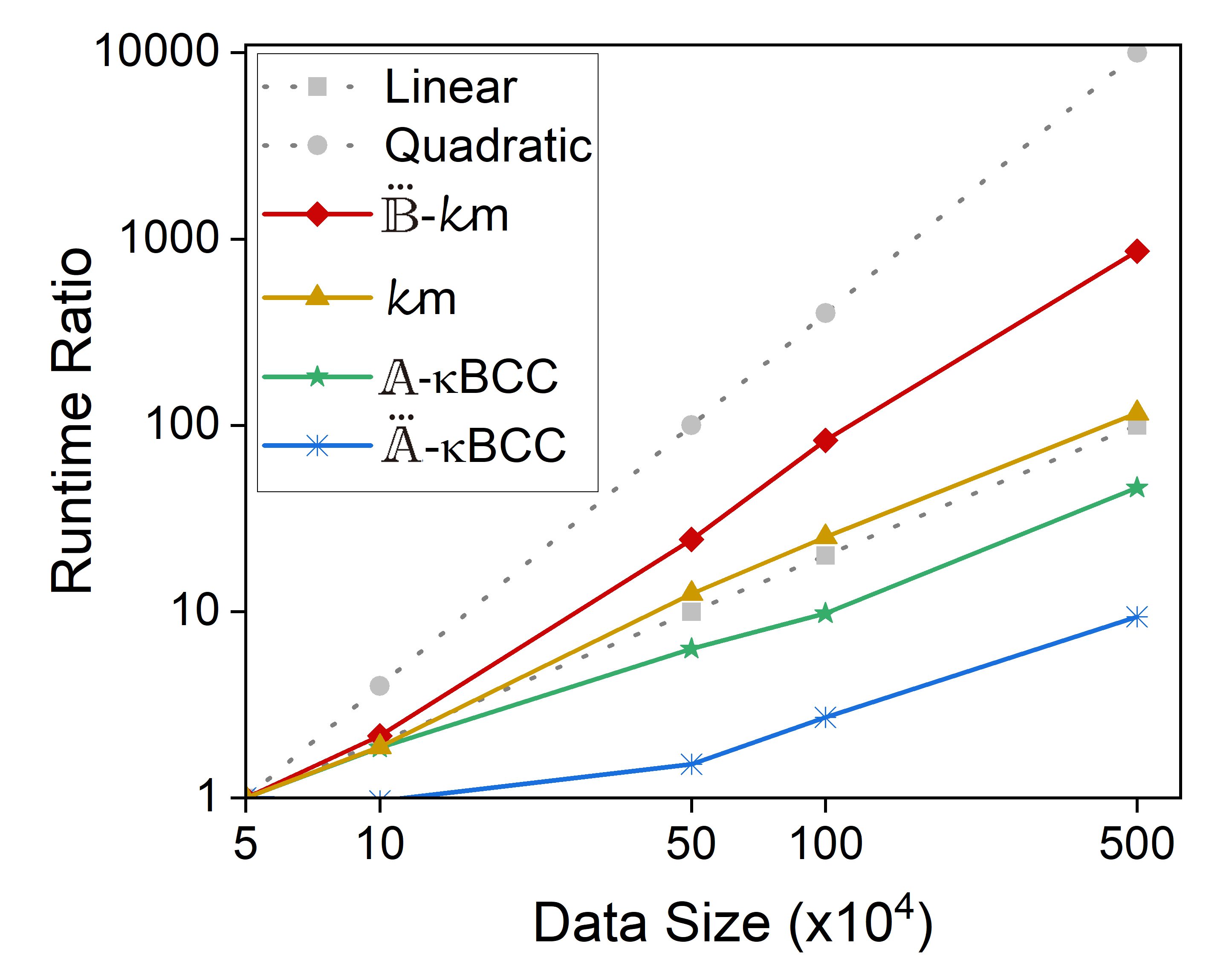}
    \caption{Scaleup test on the  mnist8m dataset. 
    }
    \label{fig:scaleup_test}
\end{figure}

\textbf{Scaleup Test}. A scaleup test comparing  Frameworks $\dddot{\mathbbm{A}}$ and $\dddot{\mathbbm{B}}$  is shown in Figure \ref{fig:scaleup_test}. The data size is increased from 50,000 to 5,000,000, sampled from the mnist8m dataset\footnote{The mnist8m dataset is available at \url{https://archive.ics.uci.edu/}.}.

We have the following observations:
\begin{enumerate}[i.]
    \item Both the proposed centralized and distributed frameworks ($\mathbbm{A}$-$\kappa$BCC and ${\dddot{\mathbbm{A}}}$-$\kappa$BCC) are sub-linear; and $\dddot{\mathbbm{A}}$-$\kappa$BCC is one order of magnitude faster than $\mathbbm{A}$-$\kappa$BCC. 
    \item The existing centralized $k$-means ($k$m) has linear time complexity\footnote{Note that $k$-means has linear time only because the maximum number of iterations is set to 100 in our experiments. Its runtime is expected to be much worse than linear if no limit is set on the maximum number of iterations.}, and the $k$-means distributed framework (${\dddot{\mathbbm{B}}}$-$k$m) has close to 
    quadratic time complexity. Note that the distributed clustering is one order of magnitude slower than the centralized version. This does not satisfy property (b) and it goes against the aim of performing distributed clustering.
    \item $\mathbbm{A}$-$\kappa$BCC and ${\dddot{\mathbbm{A}}}$-$\kappa$BCC, $k$-means and ${\dddot{\mathbbm{B}}}$-$k$m spent 480, 36, 1405 and 26813 seconds, respectively, on the dataset having 5 million points. 
\end{enumerate}

\subsection{$\mathcal{K}$DC Enables a Quadratic Time Clustering Algorithm to Deal with Large Datasets}
\label{sec:Ex-dpc++}

Here we show that ${\dddot{\mathbbm{A}}}$ can be used as a general framework that enables a quadratic time clustering algorithm to deal with datasets, that would otherwise be impossible. 

DBSCAN and Density Peak (DP) are the two most famous density-based algorithms that can find arbitrary-shaped clusters. We compare with DP here because DBSCAN often performs worse than DP \cite{IS2022Zhu, EXDPCSIGMOD,EXDPCTKDD}.

A recent DP parallel clustering algorithm is Ex-DPC++\footnote{Source codes: \url{https://github.com/amgt-d1/Ex-DPC-plus-plus}.}\cite{EXDPCSIGMOD,EXDPCTKDD}, which reduces the time complexity of DP to sub-quadratic ($O(n^{2-1/d}+n^{1.5}logn)$), More importantly, Ex-DPC++ is an exact algorithm that has some performance guarantee.
%optimal or suboptimal partitioning results. 

We employ DP \cite{DP2014} as $\mathbbm{f}$ in step 2 in ${\dddot{\mathbbm{A}}}$, where DP is a quadratic time centralized clustering algorithm. We call the resultant distributed version of DP as ${\dddot{\mathbbm{A}}}$-DP. 

\begin{figure}[t]
    \centering
    \includegraphics[width=0.7\textwidth]{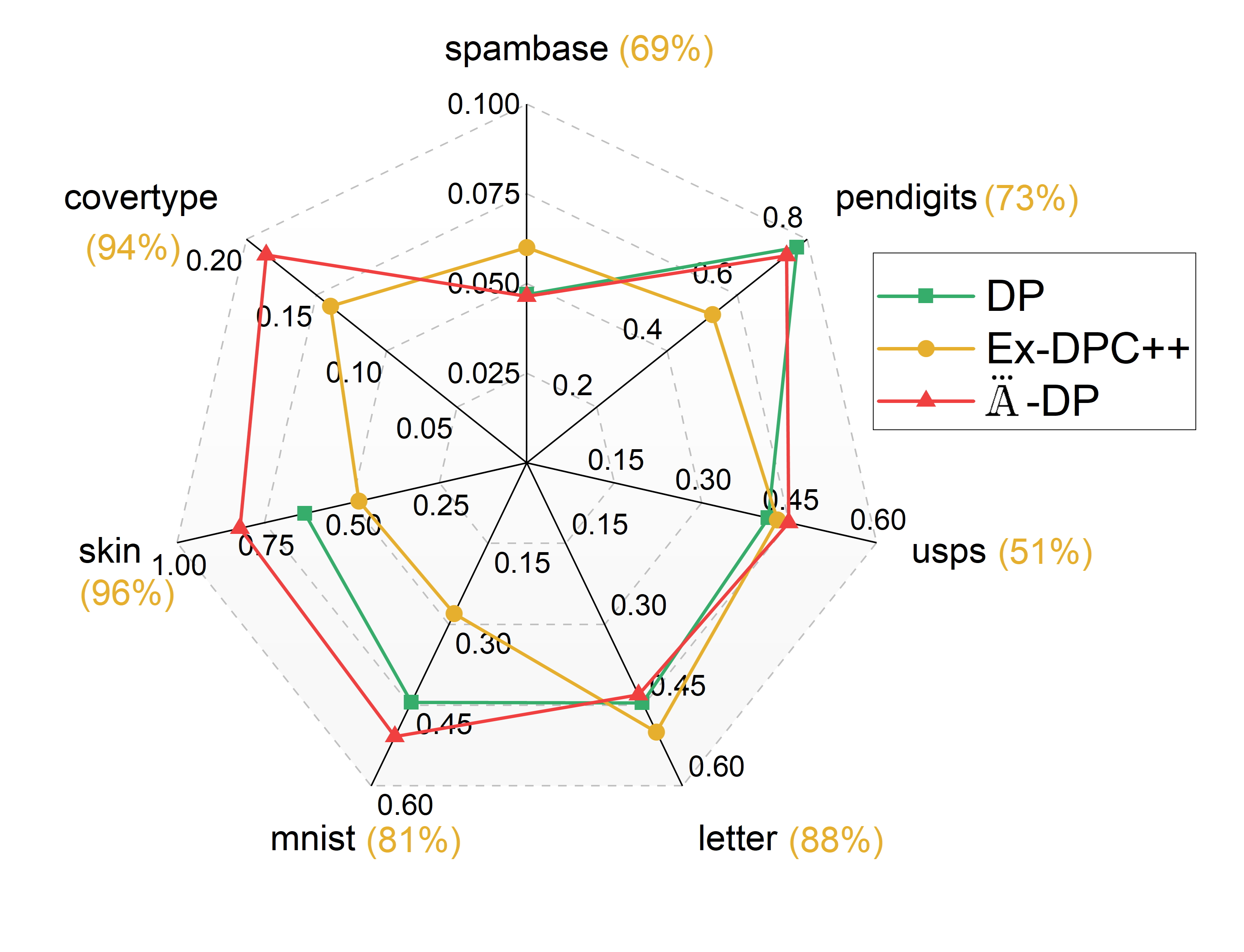}
    \caption{The comparison of standard DP and ${\dddot{\mathbbm{A}}}$-DP in terms of NMI on seven datasets. Ex-DPC++ identifies a portion of a dataset as noise. On the seven data sets, an average of 21\% of the points are identified as noise. 
    %These points are excluded when calculating NMI. 
    The percentage next to the name of each dataset denotes the percentage of points clustered by 
 Ex-DPC++ (only these points are used to calculate NMI). Both ${\dddot{\mathbbm{A}}}$-DP and DP cluster all points in each dataset.}
    \label{fig:DPd_vs_o}
\end{figure}

The result of the comparison with DP, Ex-DPC++ and ${\dddot{\mathbbm{A}}}$-DP is shown in Figure \ref{fig:DPd_vs_o}.
${\dddot{\mathbbm{A}}}$-DP approximates the clustering outcomes of DP pretty well; and ${\dddot{\mathbbm{A}}}$-DP outperforms Ex-DPC++. The average NMI  of DP, Ex-DPC++ and ${\dddot{\mathbbm{A}}}$-DP are 0.46, 0.35 and 0.50 (over six datasets excluding covertype), respectively. Note that Ex-DPC++ has the unfair advantage of clustering less than an average of 80\% of the points. Even with this advantage, it still performs substantially worse than DP (which clusters 100\% of the points) on three datasets. In short, like all methods which trade off clustering quality for efficiency, Ex-DPC++ performs worse than DP, when both cluster all points in a given dataset. On some datasets (e.g., pendigits, mnist, skin), the performance gaps are large.

On the largest covertype dataset with more than half a million points, DP could not complete the parameter search within 2 days. Yet, ${\dddot{\mathbbm{A}}}$-DP took 
%4817 seconds to complete the parameter search and took 
34 seconds to run on the optimal parameters, while Ex-DPC++ took 147 seconds.

It is instructive to compare ${\dddot{\mathbbm{A}}}$-DP with ${\dddot{\mathbbm{A}}}$-$\kappa$BCC (shown in Figure \ref{fig:main_ik}), which have average NMI of 0.45 and 0.58 (over seven datasets), respectively. This shows that $\kappa$BCC is a better clustering algorithm than DP in Framework ${\dddot{\mathbbm{A}}}$.

\section{Relative Performance of Centralized Clustering Algorithms}
\label{sec-centralized-comparison}

Every method of distributed clustering aims to achieve the clustering outcome of its centralized counterpart. As we have analyzed above in Section \ref{sec-equivalence} and Table \ref{tab:compare}, only two methods can achieve this aim, i.e., the proposed Framework $\mathbbm{A}$-$\kappa$BCC and the existing  Framework $\mathbbm{B}$ which employs coreset \cite{NIPS2013_Balcan}.
Thus, it is important to know the performance of a centralized clustering algorithm before attempting to create its distributed counterpart. An algorithm that produces a poor clustering outcome has little practical value. 

Here we compare the centralized version of $\kappa$BCC with the centralized algorithms in which distributed clustering methods have been created in the past (mentioned in Section \ref{sec_survey}). They are $k$-means, GMM \cite{gmm2015} and DP \cite{DP2014}.

A summary of the comparison of four centralized clustering algorithms in terms of NMI on 20 benchmark datasets\footnote{The details are shown at \url{https://anonymous.4open.science/r/KDC-kbcc/}. The datasets are available at \url{https://archive.ics.uci.edu/} and   \url{https://www.csie.ntu.edu.tw/$\sim$cjlin/libsvmtools/datasets/}. The biggest 3 artificial datasets are from U-SPEC \cite{USPEC}.} is shown in Table \ref{table:centralized-comparision}. The result shows that $\mathbbm{A}$-$\kappa$BCC outperforms the other three centralized clustering algorithms on almost all the datasets, where many differences are on large margins, especially in comparison with $k$-means. These results are consistent with those reported in Sections \ref{sec-exp1} and \ref{sec:Ex-dpc++}.

DP is a strong clustering algorithm which is the best among the three existing algorithms shown in Table \ref{table:centralized-comparision}. This is often attributed to its ability to find clusters of arbitrary shapes and sizes. However, our result shows that it is still worse than $\mathbbm{A}$-$\kappa$BCC. This is because DP has its own weakness for some types of clusters (see \cite{IS2022Zhu} for details). In other words, DP is unable to discover some types of clusters of arbitrary shapes, sizes and densities which can be found by $\mathbbm{A}$-$\kappa$BCC. And DP is the only algorithm that takes more than two days to run on the dataset containing 20 million points.

\begin{table}[t]
\centering
\caption{Summary Results.}
\label{table:centralized-comparision}
\scalebox{1}{
\begin{tabular}{l|ccc|cccc}
\hline
\multicolumn{4}{l|}{Algorithm}  & k-means & GMM         & DP          & $\mathbbm{A}$-$\kappa$BCC         \\ \hline
\multicolumn{4}{l|}{Average NMI}                 & 0.541   & 0.522   & 0.653  & 0.790       \\
\multicolumn{4}{l|}{Average AMI}                 & 0.530  & 0.514  & 0.637  & 0.787    \\
\multicolumn{4}{l|}{Average F1}                 & 0.541   & 0.489   & 0.618 & 0.855    \\
\multicolumn{4}{l|}{Average ARI}                 & 0.416  & 0.391   & 0.570  & 0.767       \\
\multicolumn{4}{l|}{Average Rank}                & 3.63        & 2.84        & 2.34        & 1.19       \\ 
\hline
\end{tabular}
}
\end{table}

\begin{figure}[h]
    \centering
    \includegraphics[width=0.45\textwidth]{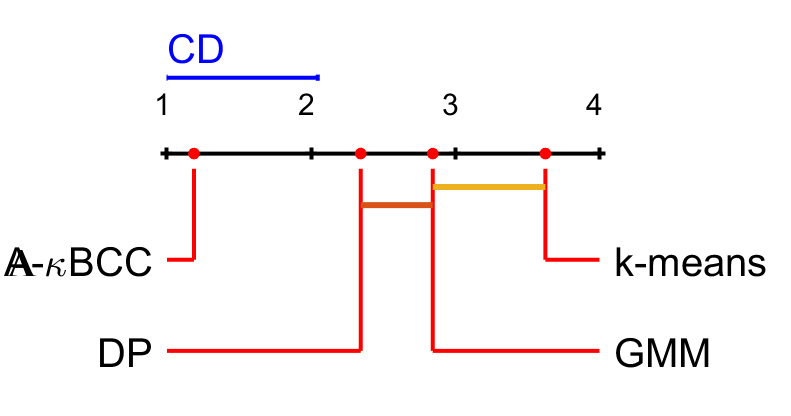}
    \caption{Nemenyi test with $\alpha$=0.1.}
    \label{fig:nemenyi_test}
\end{figure}

% \vspace{-0.8cm}
The significance test, shown in Figure \ref{fig:nemenyi_test}, reveals that $\mathbbm{A}$-$\kappa$BCC is significantly better than DP, GMM and $k$-means. We recommend that one should choose a good performing centralized clustering algorithm to produce its distributed version. 

\section{Discussion}

\subsection{Relation to Kernel $k$-means}

Kernel $k$-means clustering is an elegant way to enable $k$-means clustering to find clusters of arbitrary shapes. However, the time complexity is increased substantially to quadratic \cite{wang2019scalable}.

One way to produce a distributed version of kernel $k$-means clustering has been suggested \cite{wang2019scalable,Nystrom-2017} via the Nystr\"{o}m approximation and dimensionality reduction in order to find a low dimensional feature space. Each of these two processes cannot be performed in parallel. Once the dataset has been represented in the low dimensional feature space, $k$-means clustering is performed on the entire dataset. Thus, it is not a distributed clustering in the same sense that we have discussed so far\footnote{Although it is possible to apply Framework ${\dddot{\mathbbm{B}}}$ in its last step, the first two processes require a supercomputer to achieve the advantage of parallelization expected (see \cite{wang2019scalable,psKC-2020} for details).}. In short, distributed clustering (in the true sense) for kernel $k$-means is still an open problem.  

It is possible to view the centers in kernel $k$-means as a kind of the kernel mean maps (as used in the proposed clustering) because each cluster center is defined as the average position of all points in a cluster in the feature map. However, this interpretation does not have the concept of distribution---a richer representation of a cluster than a point in some representation. Kernel $k$-means clustering has never been regarded as distribution-based clustering (see e.g. \cite{wang2019scalable,marin2017kernel}).

The key problem with kernel $k$-means is not efficiency, but its poor clustering outcomes, compared with density-based clustering such as DP \cite{psKC-2020}. This result is consistent with ours, presented in Figure \ref{fig:main_ik} in Section \ref{sec-exp1}.

\subsection{$\dddot{\mathbbm{A}}$ is not an Extension of $\dddot{\mathbbm{B}}$}

Our proposed Framework $\dddot{\mathbbm{A}}$  is not an extension of Framework $\dddot{\mathbbm{B}}$ for three reasons. First, $\dddot{\mathbbm{B}}$ is specific to $k$-means clustering only; but the proposed $\dddot{\mathbbm{A}}$ is a generic framework that is applicable to any clustering algorithm. Second,  Framework $\dddot{\mathbbm{B}}$ applies center-based point assignment in step 3, but Framework $\dddot{\mathbbm{A}}$ applies a more powerful distribution-based point assignment. Third, Framework $\dddot{\mathbbm{A}}$ has none of the three fundamental limitations of Framework $\dddot{\mathbbm{B}}$, as stated in Section \ref{sec:intro}.

No amount of modifications to Framework $\dddot{\mathbbm{B}}$ could rectify its fundamental limitations, as a result of using $k$-means clustering. 
The proposed Framework $\dddot{\mathbbm{A}}$ is applicable to a much wider application scope than the two existing approaches, not just Framework $\dddot{\mathbbm{B}}$, because it has none of the limitations of these two approaches (stated in Section \ref{sec_survey}).

\subsection{The Impact of Unbalanced Data Sizes on Local Sites}

Many methods of distributed clustering work only if the data sizes at local sites are approximately the same. Otherwise, the clustering outcomes and/or the runtime saving is severely impacted.

\textbf{Impact on clustering outcomes}. 

This impact is well documented. Two examples are given below:
\begin{itemize}
    \item LDSDC \cite{lds20} provides the relationship between its algorithm and the number of sites. The algorithm is sensitive to the number of  sites, and the quality of the clustering outcome  degrades as the number  of sites increases.  

    \item Both DBDC and S-DBDC \cite{DBDC-2004,SDBDC-2004} usually have difficulty obtaining satisfactory parameter settings when the data sizes are not balanced over all sites.
\end{itemize}

\textbf{Impact on runtime}. The methods which do not satisfy property (b) increase their runtime significantly due to the unbalanced data sizes at different sites.

\begin{figure}[t]
    \centering
    \setlength{\abovecaptionskip}{0.cm}
    \setlength{\belowcaptionskip}{0.cm}
    \includegraphics[width=0.8\textwidth]{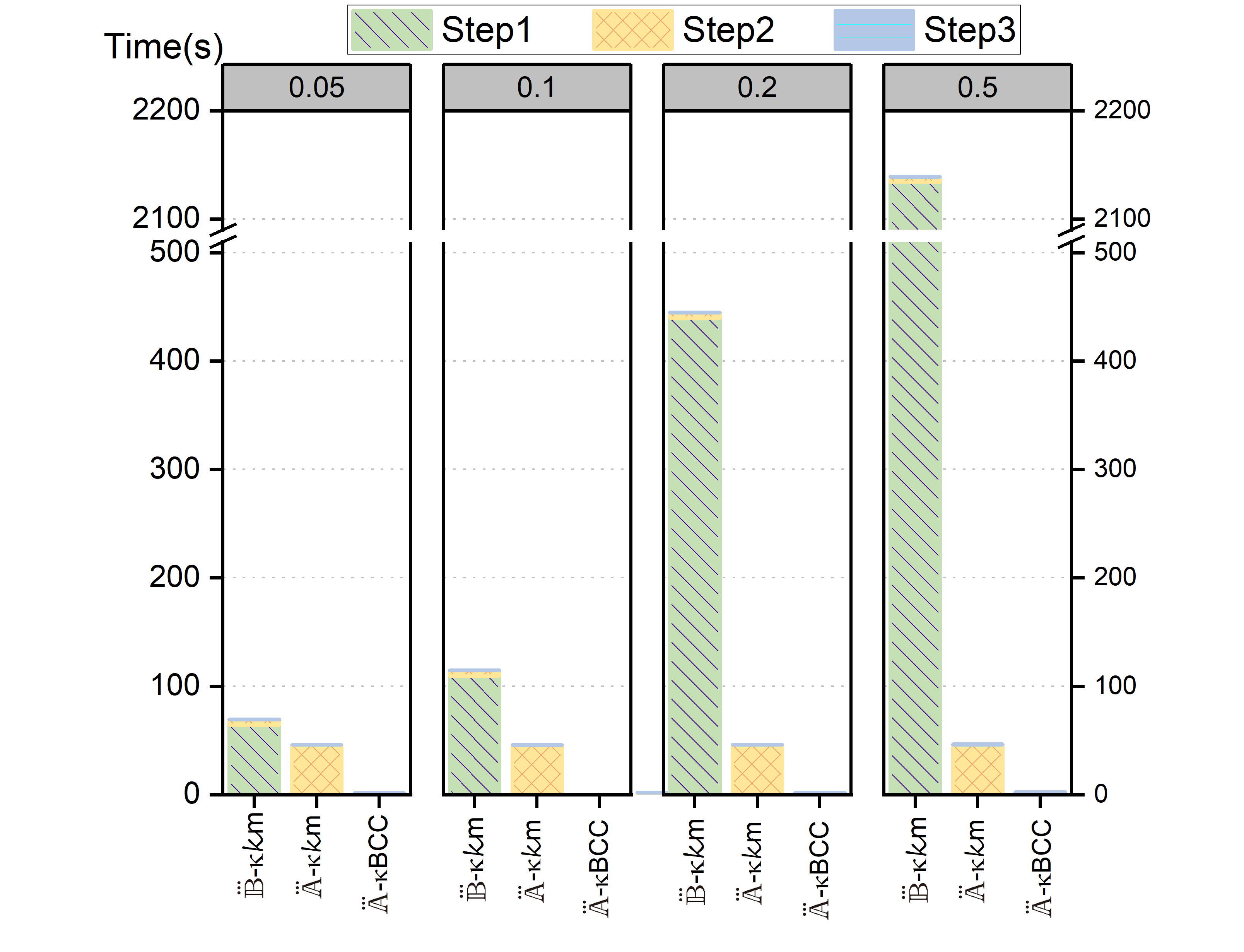}
    \caption{Runtime of the three clustering methods with different proportions (0.05, 0.1, 0.2, 0.5) of the total data size (of 14,000 points) located on the largest site in a network of $r$=20 sites. The mnist dataset is used here.}
    \label{fig:e3}
\end{figure}

Our evaluation result is shown in Figure \ref{fig:e3}. The runtimes of  ${\dddot{\mathbbm{A}}}$-$\kappa$BCC and ${\dddot{\mathbbm{A}}}$-$\kappa k$m, which satisfy property (b), are not impacted by the changing data sizes. But the runtime of ${\dddot{\mathbbm{B}}}$-$\kappa k$m, which does not satisfy property (b), increases significantly as the data size increases from 0.1 to 0.2 and 0.5 of the total data size. LDSDC \cite{lds20} and LSH-DDP \cite{ddp16} are impacted in the same way.

Although data sizes in distributed sites are assumed to be uniformly distributed in many studies \cite{ddp16,lds20,NEURIPS2018_Li,ipca14}, this scenario cannot be guaranteed in a real-world setting. Differences in the storage size are common place in the real world.

% \subsection{Intuition of ${\dddot{\mathbbm{A}}}$-$\kappa$BCC}

% Intuitively, $\mathcal{K}$DC first samples a subset from the original data, the distribution of which is very close to the distribution of the original data. Then $\kappa$BCC is used to find $\mathcal{G}$, which contains chains of core points with very high similarity. These points are often in the high-density regions of each cluster ($P_\mathcal{G}$ in Figure \ref{fig:illus}), and then the points are assigned according to the similarity with the distribution of these high-density areas.

\section{Concluding Remarks}

Current methods of distributed clustering focus on distributed computing of an \emph{existing} centralized clustering, and pay little attention on its clustering quality, knowing that the best they can achieve is to approximate the clustering outcome of the centralized clustering. 

In contrast, we emphasize on a clustering outcome which produces clusters of arbitrary shapes, sizes and densities, and 
%the distributed computing is a by-product of the proposed centralized clustering---
we design a \emph{new} clustering which is native to both centralized clustering (${\mathbbm{A}}$) and 
distributed clustering named $\mathcal{K}$DC ($\dddot{\mathbbm{A}}$). 

$\mathcal{K}$DC makes three breakthroughs in distributed clustering. First,  it is the first linear-time and distributional kernel $\mathcal{K}$ based clustering that has three properties. Out of many existing methods of distributed clustering, only the coreset-based framework possesses one out of the three properties.    

Second, the proposed use of $\mathcal{K}$ in step 3 and the proposed clustering algorithm $\kappa$BCC in step 2 of the 3-step framework contribute directly to the improved clustering outcomes in comparison with existing methods. The margin of improvement is large and significant.

Third, $\mathcal{K}$DC is the only generic framework that directly enables any quadratic-time clustering algorithm to deal with large datasets. Existing approaches are tailored made for a specific clustering algorithm only. $\mathcal{K}$DC has the ability to incorporate any clustering algorithm because it requires no parallelization of a clustering algorithm, unlike the second approach (which requires parallelization) and the first approach (which tailored for $k$-means only though requiring no parallelization) mentioned in Section \ref{sec_survey}.


% \begin{acks}

% \end{acks}


% \newpage
\bibliographystyle{ACM-Reference-Format}
\bibliography{TKDD.bib}


\begin{thebibliography}{46}

%%% the Web address.

\ifx \showCODEN    \undefined \def \showCODEN     #1{\unskip}     \fi
\ifx \showDOI      \undefined \def \showDOI       #1{#1}\fi
\ifx \showISBNx    \undefined \def \showISBNx     #1{\unskip}     \fi
\ifx \showISBNxiii \undefined \def \showISBNxiii  #1{\unskip}     \fi
\ifx \showISSN     \undefined \def \showISSN      #1{\unskip}     \fi
\ifx \showLCCN     \undefined \def \showLCCN      #1{\unskip}     \fi
\ifx \shownote     \undefined \def \shownote      #1{#1}          \fi
\ifx \showarticletitle \undefined \def \showarticletitle #1{#1}   \fi
\ifx \showURL      \undefined \def \showURL       {\relax}        \fi
% invisible to TeX
\providecommand\bibfield[2]{#2}
\providecommand\bibinfo[2]{#2}
\providecommand\natexlab[1]{#1}
\providecommand\showeprint[2][]{arXiv:#2}

\bibitem[Amagata and Hara(2021)]%
        {EXDPCSIGMOD}
\bibfield{author}{\bibinfo{person}{Daichi Amagata} {and} \bibinfo{person}{Takahiro Hara}.} \bibinfo{year}{2021}\natexlab{}.
\newblock \showarticletitle{Fast density-peaks clustering: multicore-based parallelization approach}. In \bibinfo{booktitle}{\emph{Proceedings of the 2021 ACM SIGMOD International Conference on Management of Data}}. \bibinfo{pages}{49--61}.
\newblock


\bibitem[Amagata and Hara(2023)]%
        {EXDPCTKDD}
\bibfield{author}{\bibinfo{person}{Daichi Amagata} {and} \bibinfo{person}{Takahiro Hara}.} \bibinfo{year}{2023}\natexlab{}.
\newblock \showarticletitle{Efficient Density-peaks Clustering Algorithms on Static and Dynamic Data in {E}uclidean Space}.
\newblock \bibinfo{journal}{\emph{ACM Transactions on Knowledge Discovery from Data}} \bibinfo{volume}{18}, \bibinfo{number}{1} (\bibinfo{year}{2023}), \bibinfo{pages}{1--27}.
\newblock


\bibitem[Arthur and Vassilvitskii(2006)]%
        {k-means-SCG2006}
\bibfield{author}{\bibinfo{person}{David Arthur} {and} \bibinfo{person}{Sergei Vassilvitskii}.} \bibinfo{year}{2006}\natexlab{}.
\newblock \showarticletitle{How Slow Is the k-means Method?}. In \bibinfo{booktitle}{\emph{Proceedings of the Twenty-second Annual Symposium on Computational Geometry}}. \bibinfo{pages}{144--153}.
\newblock


\bibitem[Bahmani et~al\mbox{.}(2012)]%
        {scalablekmeans}
\bibfield{author}{\bibinfo{person}{Bahman Bahmani}, \bibinfo{person}{Benjamin Moseley}, \bibinfo{person}{Andrea Vattani}, \bibinfo{person}{Ravi Kumar}, {and} \bibinfo{person}{Sergei Vassilvitskii}.} \bibinfo{year}{2012}\natexlab{}.
\newblock \showarticletitle{Scalable k-means++}.
\newblock \bibinfo{journal}{\emph{Proceedings of the VLDB Endowment}} \bibinfo{volume}{5}, \bibinfo{number}{7} (\bibinfo{year}{2012}), \bibinfo{pages}{622--633}.
\newblock


\bibitem[Balcan et~al\mbox{.}(2013a)]%
        {Coreset-kmeans-2013}
\bibfield{author}{\bibinfo{person}{M. Balcan}, \bibinfo{person}{S. Ehrlich}, {and} \bibinfo{person}{Y. Liang}.} \bibinfo{year}{2013}\natexlab{a}.
\newblock \showarticletitle{Distributed k-means and k-median Clustering on General Communication Topologies}. In \bibinfo{booktitle}{\emph{Advances in Neural Information Processing Systems}}. \bibinfo{pages}{1995--2003}.
\newblock


\bibitem[Balcan et~al\mbox{.}(2013b)]%
        {NIPS2013_Balcan}
\bibfield{author}{\bibinfo{person}{Maria-Florina~F Balcan}, \bibinfo{person}{Steven Ehrlich}, {and} \bibinfo{person}{Yingyu Liang}.} \bibinfo{year}{2013}\natexlab{b}.
\newblock \showarticletitle{Distributed k-means and k-median Clustering on General Topologies}. In \bibinfo{booktitle}{\emph{Advances in Neural Information Processing Systems}}, \bibfield{editor}{\bibinfo{person}{C.J. Burges}, \bibinfo{person}{L.~Bottou}, \bibinfo{person}{M.~Welling}, \bibinfo{person}{Z.~Ghahramani}, {and} \bibinfo{person}{K.Q. Weinberger}} (Eds.), Vol.~\bibinfo{volume}{26}. \bibinfo{publisher}{Curran Associates, Inc.}
\newblock


\bibitem[Bhaskara and Wijewardena(2018)]%
        {lsh18}
\bibfield{author}{\bibinfo{person}{Aditya Bhaskara} {and} \bibinfo{person}{Maheshakya Wijewardena}.} \bibinfo{year}{2018}\natexlab{}.
\newblock \showarticletitle{Distributed Clustering via {LSH} Based Data Partitioning}. In \bibinfo{booktitle}{\emph{Proceedings of the 35th International Conference on Machine Learning}}. \bibinfo{pages}{570--579}.
\newblock


\bibitem[Chen et~al\mbox{.}(2016)]%
        {CODC-2016}
\bibfield{author}{\bibinfo{person}{J. Chen}, \bibinfo{person}{H. Sun}, \bibinfo{person}{D.~P. Woodruff}, {and} \bibinfo{person}{Q. Zhang}.} \bibinfo{year}{2016}\natexlab{}.
\newblock \showarticletitle{Communication-optimal Distributed Clustering}. In \bibinfo{booktitle}{\emph{Advances in Neural Information Processing Systems}}. \bibinfo{pages}{3720--3728}.
\newblock


\bibitem[Christen et~al\mbox{.}(2023)]%
        {christen2023review}
\bibfield{author}{\bibinfo{person}{Peter Christen}, \bibinfo{person}{David~J Hand}, {and} \bibinfo{person}{Nishadi Kirielle}.} \bibinfo{year}{2023}\natexlab{}.
\newblock \showarticletitle{A review of the F-measure: its history, properties, criticism, and alternatives}.
\newblock \bibinfo{journal}{\emph{Comput. Surveys}} \bibinfo{volume}{56}, \bibinfo{number}{3} (\bibinfo{year}{2023}), \bibinfo{pages}{1--24}.
\newblock


\bibitem[Cohen et~al\mbox{.}(2015)]%
        {DR-kmeans-2015}
\bibfield{author}{\bibinfo{person}{M.~B. Cohen}, \bibinfo{person}{S. Elder}, \bibinfo{person}{C. Musco}, \bibinfo{person}{C. Musco}, {and} \bibinfo{person}{M. Persu}.} \bibinfo{year}{2015}\natexlab{}.
\newblock \showarticletitle{Dimensionality Reduction for k-means Clustering and Low Rank Approximation}. In \bibinfo{booktitle}{\emph{Symposium on Theory of Computing}}. \bibinfo{pages}{163--172}.
\newblock


\bibitem[Datar et~al\mbox{.}(2004)]%
        {lsh04}
\bibfield{author}{\bibinfo{person}{Mayur Datar}, \bibinfo{person}{Nicole Immorlica}, \bibinfo{person}{Piotr Indyk}, {and} \bibinfo{person}{Vahab S.Mirrokni}.} \bibinfo{year}{2004}\natexlab{}.
\newblock \showarticletitle{Locality-sensitive Hashing Scheme Based on p-stable Distributions}. In \bibinfo{booktitle}{\emph{Proceedings of the Twentieth Annual Symposium on Computational Geometry}}. \bibinfo{pages}{253–262}.
\newblock


\bibitem[Dhillon et~al\mbox{.}(2005)]%
        {complexity_kkm}
\bibfield{author}{\bibinfo{person}{Inderjit~S. Dhillon}, \bibinfo{person}{Yuqiang Guan}, {and} \bibinfo{person}{Brian Kulis}.} \bibinfo{year}{2005}\natexlab{}.
\newblock \bibinfo{booktitle}{\emph{A Unified View of Kernel k-means, Spectral Clustering and Graph Cuts}}.
\newblock \bibinfo{type}{{T}echnical {R}eport} TR-04-25. \bibinfo{institution}{University of Texas Dept. of Computer Science}.
\newblock


\bibitem[Ene et~al\mbox{.}(2011)]%
        {Map-kCenter-2011}
\bibfield{author}{\bibinfo{person}{Alina Ene}, \bibinfo{person}{Sungjin Im}, {and} \bibinfo{person}{Benjamin Moseley}.} \bibinfo{year}{2011}\natexlab{}.
\newblock \showarticletitle{Fast clustering using {M}ap{R}educe}. In \bibinfo{booktitle}{\emph{Proceedings of the 17th ACM SIGKDD international conference on Knowledge discovery and data mining}}. \bibinfo{pages}{681--689}.
\newblock


\bibitem[Ester et~al\mbox{.}(1996)]%
        {DB1996}
\bibfield{author}{\bibinfo{person}{Martin Ester}, \bibinfo{person}{Hans-Peter Kriegel}, \bibinfo{person}{Jörg Sander}, {and} \bibinfo{person}{Xiaowei Xu}.} \bibinfo{year}{1996}\natexlab{}.
\newblock \showarticletitle{A Density-based Algorithm for Discovering Clusters in Large Spatial Databases with Noise}. In \bibinfo{booktitle}{\emph{Proceedings of the Second International Conference on Knowledge Discovery and Data Mining}}. \bibinfo{pages}{226--231}.
\newblock


\bibitem[Feldman and Langberg(2011)]%
        {coreset-time}
\bibfield{author}{\bibinfo{person}{Dan Feldman} {and} \bibinfo{person}{Michael Langberg}.} \bibinfo{year}{2011}\natexlab{}.
\newblock \showarticletitle{A unified Framework for Approximating and Clustering Data}. In \bibinfo{booktitle}{\emph{Proceedings of the Forty-third Annual ACM Symposium on Theory of Computing}}. \bibinfo{pages}{569--578}.
\newblock


\bibitem[Fritz et~al\mbox{.}(2022)]%
        {fritz2022efficient}
\bibfield{author}{\bibinfo{person}{Manuel Fritz}, \bibinfo{person}{Michael Behringer}, \bibinfo{person}{Dennis Tschechlov}, {and} \bibinfo{person}{Holger Schwarz}.} \bibinfo{year}{2022}\natexlab{}.
\newblock \showarticletitle{Efficient exploratory clustering analyses in large-scale exploration processes}.
\newblock \bibinfo{journal}{\emph{The VLDB Journal}} \bibinfo{volume}{31}, \bibinfo{number}{4} (\bibinfo{year}{2022}), \bibinfo{pages}{711--732}.
\newblock


\bibitem[Geng et~al\mbox{.}(2020)]%
        {lds20}
\bibfield{author}{\bibinfo{person}{Yangli Geng}, \bibinfo{person}{Qingyong Li}, \bibinfo{person}{Mingfei Liang}, \bibinfo{person}{Chong-Yung Chi}, \bibinfo{person}{Juan Tan}, {and} \bibinfo{person}{Heng Huang}.} \bibinfo{year}{2020}\natexlab{}.
\newblock \showarticletitle{Local-Density Subspace Distributed Clustering for High-Dimensional Data}.
\newblock \bibinfo{journal}{\emph{IEEE Transactions on Parallel and Distributed System}} \bibinfo{volume}{31}, \bibinfo{number}{8} (\bibinfo{year}{2020}), \bibinfo{pages}{1799--1814}.
\newblock


\bibitem[Guo and Li(2018)]%
        {NEURIPS2018_Li}
\bibfield{author}{\bibinfo{person}{Xiangyu Guo} {and} \bibinfo{person}{Shi Li}.} \bibinfo{year}{2018}\natexlab{}.
\newblock \showarticletitle{Distributed k-clustering for data with heavy noise}. In \bibinfo{booktitle}{\emph{Proceedings of the 32nd International Conference on Neural Information Processing Systems}}. \bibinfo{pages}{7849--7857}.
\newblock


\bibitem[Hartigan and Wong(1979)]%
        {hartigan1979algorithm}
\bibfield{author}{\bibinfo{person}{John~A Hartigan} {and} \bibinfo{person}{Manchek~A Wong}.} \bibinfo{year}{1979}\natexlab{}.
\newblock \showarticletitle{Algorithm AS 136: A k-means clustering algorithm}.
\newblock \bibinfo{journal}{\emph{Journal of the Royal Statistical Society. Series c (Applied Statistics)}} \bibinfo{volume}{28}, \bibinfo{number}{1} (\bibinfo{year}{1979}), \bibinfo{pages}{100--108}.
\newblock


\bibitem[Huang et~al\mbox{.}(2020)]%
        {USPEC}
\bibfield{author}{\bibinfo{person}{Dong Huang}, \bibinfo{person}{Chang-Dong Wang}, \bibinfo{person}{Jian-Sheng Wu}, \bibinfo{person}{Jian-Huang Lai}, {and} \bibinfo{person}{Chee-Keong Kwoh}.} \bibinfo{year}{2020}\natexlab{}.
\newblock \showarticletitle{Ultra-Scalable Spectral Clustering and Ensemble Clustering}.
\newblock \bibinfo{journal}{\emph{IEEE Transactions on Knowledge and Data Engineering}} \bibinfo{volume}{32}, \bibinfo{number}{6} (\bibinfo{year}{2020}), \bibinfo{pages}{1212--1226}.
\newblock


\bibitem[Januzaj et~al\mbox{.}(2004a)]%
        {DBDC-2004}
\bibfield{author}{\bibinfo{person}{E. Januzaj}, \bibinfo{person}{H.-P. Kriegel}, {and} \bibinfo{person}{M. Pfeifle}.} \bibinfo{year}{2004}\natexlab{a}.
\newblock \showarticletitle{{DBDC}: Density-Based Distributed Clustering}. In \bibinfo{booktitle}{\emph{Proceedings of the 9th International Conference on Extending Database Technology}}. \bibinfo{pages}{88--105}.
\newblock


\bibitem[Januzaj et~al\mbox{.}(2004b)]%
        {SDBDC-2004}
\bibfield{author}{\bibinfo{person}{E. Januzaj}, \bibinfo{person}{H.-P. Kriegel}, {and} \bibinfo{person}{M. Pfeifle}.} \bibinfo{year}{2004}\natexlab{b}.
\newblock \showarticletitle{Scalable Density-Based Distributed Clustering}. In \bibinfo{booktitle}{\emph{Proceedings of the European Conference on Principles of Data Mining and Knowledge Discovery}}. \bibinfo{pages}{231--244}.
\newblock


\bibitem[Kantabutra and Couch(2000)]%
        {parakmeans2000}
\bibfield{author}{\bibinfo{person}{Sanpawat Kantabutra} {and} \bibinfo{person}{Alva~L. Couch}.} \bibinfo{year}{2000}\natexlab{}.
\newblock \showarticletitle{Parallel k-means Clustering Algorithm on Nows}.
\newblock \bibinfo{journal}{\emph{NECTEC Technical Journal}} \bibinfo{volume}{1}, \bibinfo{number}{6} (\bibinfo{year}{2000}), \bibinfo{pages}{243--247}.
\newblock


\bibitem[Liang et~al\mbox{.}(2014)]%
        {ipca14}
\bibfield{author}{\bibinfo{person}{Yingyu Liang}, \bibinfo{person}{Maria-Florina~F Balcan}, \bibinfo{person}{Vandana Kanchanapally}, {and} \bibinfo{person}{David Woodruff}.} \bibinfo{year}{2014}\natexlab{}.
\newblock \showarticletitle{Improved distributed principal component analysis}.
\newblock \bibinfo{journal}{\emph{Advances in Neural Information Processing Systems}}  \bibinfo{volume}{27} (\bibinfo{year}{2014}).
\newblock


\bibitem[Lu et~al\mbox{.}(2020)]%
        {lu2020distributed}
\bibfield{author}{\bibinfo{person}{Jing Lu}, \bibinfo{person}{Yuhai Zhao}, \bibinfo{person}{Kian-Lee Tan}, {and} \bibinfo{person}{Zhengkui Wang}.} \bibinfo{year}{2020}\natexlab{}.
\newblock \showarticletitle{Distributed density peaks clustering revisited}.
\newblock \bibinfo{journal}{\emph{IEEE Transactions on Knowledge and Data Engineering}} \bibinfo{volume}{34}, \bibinfo{number}{8} (\bibinfo{year}{2020}), \bibinfo{pages}{3714--3726}.
\newblock


\bibitem[Lulli et~al\mbox{.}(2016)]%
        {NG_DBSCAN}
\bibfield{author}{\bibinfo{person}{Alessandro Lulli}, \bibinfo{person}{Matteo Dell'Amico}, \bibinfo{person}{Pietro Michiardi}, {and} \bibinfo{person}{Laura Ricci}.} \bibinfo{year}{2016}\natexlab{}.
\newblock \showarticletitle{{NG-DBSCAN}: Scalable Density-Based Clustering for Arbitrary Data}.
\newblock \bibinfo{journal}{\emph{Proceedings of the VLDB Endowment}} \bibinfo{volume}{10}, \bibinfo{number}{3} (\bibinfo{date}{nov} \bibinfo{year}{2016}), \bibinfo{pages}{157–168}.
\newblock


\bibitem[Marin et~al\mbox{.}(2017)]%
        {marin2017kernel}
\bibfield{author}{\bibinfo{person}{Dmitrii Marin}, \bibinfo{person}{Meng Tang}, \bibinfo{person}{Ismail~Ben Ayed}, {and} \bibinfo{person}{Yuri Boykov}.} \bibinfo{year}{2017}\natexlab{}.
\newblock \showarticletitle{Kernel Clustering: Density Biases and Solutions}.
\newblock \bibinfo{journal}{\emph{IEEE Transactions on Pattern Analysis and Machine Intelligence}} \bibinfo{volume}{41}, \bibinfo{number}{1} (\bibinfo{year}{2017}), \bibinfo{pages}{136--147}.
\newblock


\bibitem[Muandet et~al\mbox{.}(2017)]%
        {KernelMeanEmbedding2017}
\bibfield{author}{\bibinfo{person}{Krikamol Muandet}, \bibinfo{person}{Kenji Fukumizu}, \bibinfo{person}{Bharath Sriperumbudur}, {and} \bibinfo{person}{Bernhard Schölkopf}.} \bibinfo{year}{2017}\natexlab{}.
\newblock \showarticletitle{Kernel Mean Embedding of Distributions: A Review and Beyond}.
\newblock \bibinfo{journal}{\emph{Foundations and Trends in Machine Learning}}  \bibinfo{volume}{10 (1–2)} (\bibinfo{year}{2017}), \bibinfo{pages}{1--141}.
\newblock


\bibitem[Musco and Musco(2017)]%
        {Nystrom-2017}
\bibfield{author}{\bibinfo{person}{Cameron Musco} {and} \bibinfo{person}{Christopher Musco}.} \bibinfo{year}{2017}\natexlab{}.
\newblock \showarticletitle{Recursive Sampling for the {N}ystr\"{o}m Method}. In \bibinfo{booktitle}{\emph{Advances in Neural Information Processing Systems}}, Vol.~\bibinfo{volume}{30}.
\newblock


\bibitem[Ni et~al\mbox{.}(2008)]%
        {LDBDC-2008}
\bibfield{author}{\bibinfo{person}{W. Ni}, \bibinfo{person}{G. Chen}, \bibinfo{person}{Y.J. Wu}, {and} \bibinfo{person}{Z.H. Sun}.} \bibinfo{year}{2008}\natexlab{}.
\newblock \showarticletitle{Local Density Based Distributed Clustering Algorithm}.
\newblock \bibinfo{journal}{\emph{Journal of Software}} \bibinfo{volume}{19}, \bibinfo{number}{9} (\bibinfo{year}{2008}), \bibinfo{pages}{2339--2348}.
\newblock


\bibitem[Ordonez and Garc{\'\i}a-Garc{\'\i}a(2010)]%
        {ordonez2010database}
\bibfield{author}{\bibinfo{person}{Carlos Ordonez} {and} \bibinfo{person}{Javier Garc{\'\i}a-Garc{\'\i}a}.} \bibinfo{year}{2010}\natexlab{}.
\newblock \showarticletitle{Database Systems Research on Data Mining}. In \bibinfo{booktitle}{\emph{Proceedings of the 2010 ACM SIGMOD International Conference on Management of Data}}. \bibinfo{pages}{1253--1254}.
\newblock


\bibitem[Reynolds(2009)]%
        {gmm2015}
\bibfield{author}{\bibinfo{person}{Douglas~A Reynolds}.} \bibinfo{year}{2009}\natexlab{}.
\newblock \showarticletitle{Gaussian mixture models.}
\newblock \bibinfo{journal}{\emph{Encyclopedia of biometrics}} \bibinfo{volume}{741}, \bibinfo{number}{659-663} (\bibinfo{year}{2009}).
\newblock


\bibitem[Rodriguez and Laio(2014)]%
        {DP2014}
\bibfield{author}{\bibinfo{person}{Alex Rodriguez} {and} \bibinfo{person}{Alessandro Laio}.} \bibinfo{year}{2014}\natexlab{}.
\newblock \showarticletitle{Clustering by Fast Search and Find of Density Peaks}.
\newblock \bibinfo{journal}{\emph{Science}}  \bibinfo{volume}{344} (\bibinfo{year}{2014}), \bibinfo{pages}{1492–--1496}.
\newblock


\bibitem[Sheikholeslami et~al\mbox{.}(2000)]%
        {wavecluster}
\bibfield{author}{\bibinfo{person}{Gholamhosein Sheikholeslami}, \bibinfo{person}{Surojit Chatterjee}, {and} \bibinfo{person}{Aidong Zhang}.} \bibinfo{year}{2000}\natexlab{}.
\newblock \showarticletitle{WaveCluster: a wavelet-based clustering approach for spatial data in very large databases}.
\newblock \bibinfo{journal}{\emph{The VLDB Journal}} \bibinfo{volume}{8}, \bibinfo{number}{3} (\bibinfo{year}{2000}), \bibinfo{pages}{289--304}.
\newblock


\bibitem[Song and Lee(2018)]%
        {RP-DBSCAN}
\bibfield{author}{\bibinfo{person}{Hwanjun Song} {and} \bibinfo{person}{Jae-Gil Lee}.} \bibinfo{year}{2018}\natexlab{}.
\newblock \showarticletitle{{RP-DBSCAN}: A Superfast Parallel DBSCAN Algorithm Based on Random Partitioning}. In \bibinfo{booktitle}{\emph{Proceedings of the 2018 ACM SIGMOD International Conference on Management of Data}}. ACM, \bibinfo{pages}{1173--1187}.
\newblock


\bibitem[Steinley(2004)]%
        {steinley2004properties}
\bibfield{author}{\bibinfo{person}{Douglas Steinley}.} \bibinfo{year}{2004}\natexlab{}.
\newblock \showarticletitle{Properties of the hubert-arable adjusted rand index.}
\newblock \bibinfo{journal}{\emph{Psychological methods}} \bibinfo{volume}{9}, \bibinfo{number}{3} (\bibinfo{year}{2004}), \bibinfo{pages}{386}.
\newblock


\bibitem[Ting et~al\mbox{.}(2022)]%
        {psKC-2020}
\bibfield{author}{\bibinfo{person}{Kai~Ming Ting}, \bibinfo{person}{Jonathan~R Wells}, {and} \bibinfo{person}{Ye Zhu}.} \bibinfo{year}{2022}\natexlab{}.
\newblock \showarticletitle{Point-Set Kernel Clustering}.
\newblock \bibinfo{journal}{\emph{IEEE Transactions on Knowledge and Data Engineering}} (\bibinfo{year}{2022}).
\newblock


\bibitem[Ting et~al\mbox{.}(2018)]%
        {ting2018IsolationKernel}
\bibfield{author}{\bibinfo{person}{Kai~Ming Ting}, \bibinfo{person}{Yue Zhu}, {and} \bibinfo{person}{Zhi-Hua Zhou}.} \bibinfo{year}{2018}\natexlab{}.
\newblock \showarticletitle{Isolation Kernel and Its Effect on {SVM}}. In \bibinfo{booktitle}{\emph{Proceedings of the 24th ACM International Conference on Knowledge Discovery and Data Mining}}. ACM, \bibinfo{pages}{2329--2337}.
\newblock


\bibitem[Vinh et~al\mbox{.}(2009)]%
        {AMI}
\bibfield{author}{\bibinfo{person}{Nguyen~Xuan Vinh}, \bibinfo{person}{Julien Epps}, {and} \bibinfo{person}{James Bailey}.} \bibinfo{year}{2009}\natexlab{}.
\newblock \showarticletitle{Information theoretic measures for clusterings comparison: is a correction for chance necessary?}. In \bibinfo{booktitle}{\emph{Proceedings of the 26th Annual International Conference on Machine Learning}} (Montreal, Quebec, Canada). \bibinfo{publisher}{Association for Computing Machinery}, \bibinfo{address}{New York, NY, USA}, \bibinfo{pages}{1073–1080}.
\newblock
\showISBNx{9781605585161}


\bibitem[Vinh et~al\mbox{.}(2010)]%
        {NMI2010}
\bibfield{author}{\bibinfo{person}{Nguyen~Xuan Vinh}, \bibinfo{person}{Julien Epps}, {and} \bibinfo{person}{James Bailey}.} \bibinfo{year}{2010}\natexlab{}.
\newblock \showarticletitle{Information Theoretic Measures for Clusterings Comparison: Variants, Properties, Normalization and Correction for Chance}.
\newblock \bibinfo{journal}{\emph{The Journal of Machine Learning Research}}  \bibinfo{volume}{11} (\bibinfo{year}{2010}), \bibinfo{pages}{2837--2854}.
\newblock


\bibitem[Wang et~al\mbox{.}(2019)]%
        {wang2019scalable}
\bibfield{author}{\bibinfo{person}{Shusen Wang}, \bibinfo{person}{Alex Gittens}, {and} \bibinfo{person}{Michael~W Mahoney}.} \bibinfo{year}{2019}\natexlab{}.
\newblock \showarticletitle{Scalable Kernel K-means Clustering with {N}ystr\"{o}m Approximation: Relative-error Bounds}.
\newblock \bibinfo{journal}{\emph{The Journal of Machine Learning Research}} \bibinfo{volume}{20}, \bibinfo{number}{1} (\bibinfo{year}{2019}), \bibinfo{pages}{431--479}.
\newblock


\bibitem[Wang et~al\mbox{.}(2020)]%
        {TEPDBSCAN}
\bibfield{author}{\bibinfo{person}{Yiqiu Wang}, \bibinfo{person}{Yan Gu}, {and} \bibinfo{person}{Julian Shun}.} \bibinfo{year}{2020}\natexlab{}.
\newblock \showarticletitle{Theoretically-Efficient and Practical Parallel DBSCAN}. In \bibinfo{booktitle}{\emph{Proceedings of the 2020 ACM SIGMOD International Conference on Management of Data}} (Portland, OR, USA). \bibinfo{publisher}{Association for Computing Machinery}, \bibinfo{pages}{2555–2571}.
\newblock


\bibitem[Wang et~al\mbox{.}(2022)]%
        {wang2022pack}
\bibfield{author}{\bibinfo{person}{Yue Wang}, \bibinfo{person}{Vivek Narasayya}, \bibinfo{person}{Yeye He}, {and} \bibinfo{person}{Surajit Chaudhuri}.} \bibinfo{year}{2022}\natexlab{}.
\newblock \showarticletitle{{PACk}: an efficient partition-based distributed agglomerative hierarchical clustering algorithm for deduplication}.
\newblock \bibinfo{journal}{\emph{Proceedings of the VLDB Endowment}} \bibinfo{volume}{15}, \bibinfo{number}{6} (\bibinfo{year}{2022}), \bibinfo{pages}{1132--1145}.
\newblock


\bibitem[Zhang et~al\mbox{.}(2016a)]%
        {zhang2016efficient}
\bibfield{author}{\bibinfo{person}{Yanfeng Zhang}, \bibinfo{person}{Shimin Chen}, {and} \bibinfo{person}{Ge Yu}.} \bibinfo{year}{2016}\natexlab{a}.
\newblock \showarticletitle{Efficient distributed density peaks for clustering large data sets in mapreduce}.
\newblock \bibinfo{journal}{\emph{IEEE Transactions on Knowledge and Data Engineering}} \bibinfo{volume}{28}, \bibinfo{number}{12} (\bibinfo{year}{2016}), \bibinfo{pages}{3218--3230}.
\newblock


\bibitem[Zhang et~al\mbox{.}(2016b)]%
        {ddp16}
\bibfield{author}{\bibinfo{person}{Yanfeng Zhang}, \bibinfo{person}{Shimin Chen}, {and} \bibinfo{person}{Ge Yu}.} \bibinfo{year}{2016}\natexlab{b}.
\newblock \showarticletitle{Efficient Distributed Density Peaks for Clustering Large Data Sets in MapReduce}.
\newblock \bibinfo{journal}{\emph{IEEE Transactions on Knowledge and Data Engineering}} \bibinfo{volume}{28}, \bibinfo{number}{12} (\bibinfo{year}{2016}), \bibinfo{pages}{3218--3230}.
\newblock


\bibitem[Zhu et~al\mbox{.}(2022)]%
        {IS2022Zhu}
\bibfield{author}{\bibinfo{person}{Ye Zhu}, \bibinfo{person}{Kai~Ming Ting}, \bibinfo{person}{Yuan Jin}, {and} \bibinfo{person}{Maia Angelova}.} \bibinfo{year}{2022}\natexlab{}.
\newblock \showarticletitle{Hierarchical clustering that takes advantage of both density-peak and density-connectivity}.
\newblock \bibinfo{journal}{\emph{Information Systems}} \bibinfo{volume}{103}, \bibinfo{number}{101871} (\bibinfo{year}{2022}).
\newblock


\end{thebibliography}

% \appendix

\end{document}